%% file: main.tex
\documentclass[10pt,twocolumn,letterpaper]{article}

\usepackage{cvpr}              % To produce the CAMERA-READY version
\input{preamble}
\definecolor{cvprblue}{rgb}{0.21,0.49,0.74}
\usepackage[pagebackref,breaklinks,colorlinks,allcolors=cvprblue]{hyperref}

\def\paperID{} % *** Enter the Paper ID here
\def\confName{CVPR}
\def\confYear{2026}

\title{Organ-Aware Attention Improves CT Triage and Classification}

\author{
Lavsen Dahal\\
Duke University\\
Durham, NC, USA\\
{\tt\small lavsen.dahal@duke.edu}
\and
Yubraj Bhandari\\
Duke University\\
Durham, NC, USA\\
{\tt\small yubraj.bhandari@duke.edu}
\and
Geoffrey D. Rubin\\
University of Arizona\\
Tucson, AZ, USA\\
{\tt\small grubin@arizona.edu}
\and
Joseph Y. Lo\\
Duke University\\
Durham, NC, USA\\
{\tt\small joseph.lo@duke.edu}
}

\begin{document}
\maketitle
\input{sec/0_abstract}    
\input{sec/1_intro}
\input{sec/2_formatting}
\input{sec/3_finalcopy}
{
    \small
    \bibliographystyle{plain}
    \bibliography{main}
}

% WARNING: do not forget to delete the supplementary pages from your submission 
\input{sec/X_suppl}

\end{document}

%% file: sec/0_abstract.tex
\begin{abstract}
There is an urgent need for triage and classification of high-volume medical imaging modalities such as computed tomography (CT), which can improve patient care and mitigate radiologist burnout. Study-level CT triage requires calibrated predictions with localized evidence; however, off-the-shelf Vision–Language Models (VLM) struggle with 3D anatomy, protocol shifts, and noisy report supervision. This study used the two largest publicly available chest CT datasets—CT-RATE and RADCHEST-CT (held-out external test set). Our carefully tuned supervised baseline (instantiated as a simple Global Average Pooling head) establishes a new supervised state of the art, surpassing all reported linear-probe VLMs. Building on this baseline, we present ORACLE-CT, an encoder-agnostic, organ-aware head that pairs Organ-Masked Attention (mask-restricted, per-organ pooling that yields spatial evidence) with Organ-Scalar Fusion (lightweight fusion of normalized volume and mean-HU cues). In the chest setting, ORACLE-CT’s masked attention model achieves AUROC 0.86 on CT-RATE; in the abdomen setting, on MERLIN (30 findings), our supervised baseline exceeds a reproduced zero-shot VLM baseline obtained by running publicly released weights through our pipeline, and adding masked attention plus scalar fusion further improves performance to AUROC 0.85. Together, these results deliver state-of-the-art supervised classification performance across both chest and abdomen CT under a unified evaluation protocol. The source code is available at https://github.com/lavsendahal/oracle-ct. 
\end{abstract}

%% file: sec/1_intro.tex
\section{Introduction}
\label{sec:intro}

Computed tomography (CT) is among the most widely used imaging modalities, with over 90M scans performed in the United States in 2023~\cite{oecd_ct_exams_indicator, smith2025projected}. Chest and abdomen–pelvis studies comprise over half of these exams, and volumes continue to rise while the radiologist workforce does not~\cite{winder2021we}, increasing backlog, and reducing time per case~\cite{vosshenrich2021quantifying, pourvaziri2022imaging}. As each CT dataset is a large, 3D volume, radiologists must interrogate hundreds of slices per study across heterogeneous scanners and protocols (non-contrast/contrast, arterial/venous phases), making the detection and classification of disease time-consuming and error-prone~\cite{kelly2015incidental, evans2022incidental, ritter2025incidental}.

At the same time, computer vision has advanced rapidly. Convolutional neural networks (CNNs) and self-supervised Vision Transformers (ViT) (\eg, ResNet, Inception, DINOv3)~\cite{he2016deep, szegedy2015going, simeoni2025dinov3, liu2025does}, have propelled image classification, while accurate multi-organ 3D segmentation is now accessible via open-source or commercial tools~\cite{wasserthal2023totalsegmentator, dahal2025xcat, Siemens_syngovia_accessed2025,Synopsys_Simpleware_AutoSeg_accessed2025}. In parallel, large language models (LLMs)~\cite{vaswani2017attention, touvron2023llama, brown2020language, Google_LangExtract_accessed2025} reduce annotation burden by parsing radiology text reports into structured findings~\cite{mukherjee2023feasibility,reichenpfader2024scoping,le2024performance, garcia2025evaluating}, enabling large-scale supervised CT training.

Vision–language models (VLM) excel on natural images via large-scale image–text contrastive pretraining, yielding transferable features and strong zero-/few-shot generalization~\cite{radford2021learning}. But directly porting this recipe to CT exposes three mismatches: (i) \textbf{modality/geometry}—3D CT volumes with quantitative, physically calibrated Hounsfield Units (HU) with protocol-dependent contrast rather than 2D RGB photos; (ii) \textbf{supervision}—radiology reports are long and heterogeneous, interleaving positive and negative findings, prior comparisons, and recommendations, vs. short object-centric captions; and (iii) \textbf{alignment}—organ and finding-level evidence is weakly grounded in text and spatially sparse/uneven across slices. Consequently, contrastive objectives that work in natural images do not automatically yield reliable calibrated clinical behavior~\cite{shui2025large,wald2025comprehensive,lin2024ct}.

In response, CT–VLMs have taken two paths: \textbf{(A) Global} models align an entire study with its full report (e.g., CT-CLIP/CT-GLIP), which is convenient but coarse and prone to \emph{implicit negatives}, reducing organ-level interpretability and stability~\cite{hamamci2024developing,lin2024ct}. They also show strong \emph{prompt sensitivity}, with zero-shot performance varying across near-equivalent phrasings—problematic for auditable triage~\cite{wald2025comprehensive}. \textbf{(B) Anatomy-aware} models (\eg fVLM) add 3D masks and organ-wise text to mitigate misalignment and lift zero-shot AUCs~\cite{shui2025large,lee2025unified}. Yet across \emph{both} families, core issues persist: sparse organ-finding-level grounding in reports, sensitivity to prompt wording and calibration choices~\cite{wald2025comprehensive,blankemeier2024merlin} that do not automatically yield reliable, well-calibrated behavior for clinical CT triage. This motivates a third path: a supervised, anatomy-guided pooling head that optimizes directly for CT labels. We adopt a \emph{Multiple Instance Learning (MIL)} formulation: each organ group mask defines a bag of token/voxel instances; a small \emph{unary} scorer assigns one scalar per instance and a \emph{mask-restricted softmax} yields a permutation-invariant organ embedding for study-level classification. It is soft attention over sets~\cite{dietterich1997solving,ilse2018attention,zaheer2017deep, bahdanau2014neural, xu2015show, luong2015effective} using segmentation masks, while avoiding pairwise Q--K--V mixing used in Transformers~\cite{vaswani2017attention,wang2018non,cheng2022masked}.

\textbf{Our approach.} We introduce \emph{ORACLE-CT} (\underline{OR}gan-\underline{A}ware \underline{C}lassification \underline{E}ngine for \underline{CT}), which has two complementary pillars. \emph{(i) Organ-masked softmax pooling (unary) with Organ-Scalar Fusion (OSF):} for each organ, a unary scorer produces one scalar per local feature; we normalize \emph{within the organ mask} and pool an organ embedding, then (optionally) append compact HU/volume scalars for classification. \emph{(ii) An encoder-agnostic classification engine and fixed, auditable protocol:} any 2D/2.5D/3D backbone that outputs a feature lattice can be plugged in, while preprocessing, label maps, splits, calibration (temperatures/thresholds fixed on validation), and decision rules are held constant. This design yields calibrated operating points and \emph{auditable} organ-weight maps; we treat these maps as supportive evidence rather than causal attributions~\cite{jain2019attention}. As \emph{ORACLE--CT} uses the same multi-organ masks as anatomy-aware VLMs, segmentation is a \emph{shared, once-per-study} cost reusable across labels and models~\cite{wasserthal2023totalsegmentator,dahal2025xcat,Synopsys_Simpleware_AutoSeg_accessed2025}. The key difference is \emph{objective alignment}: we learn directly from CT disease labels rather than proxy text alignment, reducing exposure to modality and supervision gaps. We therefore report side-by-side with CT–VLM references using their strongest linear-probe or zero-shot results under our harmonized protocol~\cite{hamamci2024developing,shui2025large,lee2025unified,lin2024ct,wald2025comprehensive}, and we stage ablations from \emph{Global Average Pooling (GAP)} to \emph{global attention (mask-free)}, \emph{organ-masked pooling}, and \emph{masked+OSF} to quantify the incremental value of anatomy guidance.

\paragraph{Contributions.}
\begin{itemize}
  \item \textbf{Organ-masked softmax pooling for CT Triage.} A lightweight, encoder-agnostic pooling head that scores each local feature with a unary scalar and \emph{normalizes within the organ mask}, yielding permutation-invariant organ embeddings and spatially localized weight maps suitable for auditing.

  \item \textbf{Organ-Scalar Fusion (OSF).} A minimal extension that appends compact, mask-derived scalars (e.g., normalized volume, mean HU, truncation flag) to the pooled organ embedding, improving numerous findings sensitive to size or density.

  \item \textbf{Encoder-agnostic benchmarking protocol.} A single modular engine compares diverse 2.5D and 3D backbones under fixed preprocessing, label maps, splits, schedules, and calibration—enabling rigorous GAP vs.\ global attention vs.\ masked attention vs.\ masked attention + OSF comparisons.

  \item \textbf{Calibrated supervised SoTA.} With fixed preprocessing, splits, and calibration, \emph{ORACLE--CT} establishes supervised SoTA on \textsc{CT--RATE} and \textsc{Merlin}, outperforming prior supervised baselines and competitive linear-probe CT–VLMs.
  
\end{itemize}

\section{Related Work}

Early supervised baselines established feasibility at scale: a single-institution study on $\sim$36k chest CTs reported AUROC $0.77$ over 83 findings and released \emph{Rad-ChestCT}~\cite{draelos2021machine}. Consistent with prior work, \emph{diffuse} disease reliably outperform \emph{focal} lesions across organs—lungs (effusion/emphysema vs.\ nodules), liver (steatosis/biliary dilation vs.\ lesions/calcifications), and kidneys (stones/atrophy vs.\ small lesions/cysts)~\cite{tushar2021classification}. \emph{CT-RATE} then introduced paired 3D CT–report supervision; its VocabFine reached AUROC $0.756$ (internal) and $0.650$ (external)~\cite{hamamci2024developing}. Anatomy-aware CT–VLMs further improved metrics: \emph{fVLM} achieved $0.780$ (internal) and $0.646$ (external)~\cite{shui2025large}, while segmentation-guided variants (e.g., SegVL) report $0.811$ (internal) and $0.716$ (external)~\cite{husegmentation}. \emph{Uniferum} unifies classification and segmentation signals and reports AUROC $0.83$ on \emph{CT-RATE}~\cite{lee2025unified}. Most recently, \emph{COLIPRI} conducted an extensive ablation of contrastive pretraining, report-generation, and masked-image modeling objectives and reported the strongest \emph{linear-probe} baselines to date: AUROC 84.15 on \emph{CT-RATE} and 72.66 on \emph{Rad-ChestCT} with frozen encoders. Notably, their zero-shot performance is highly prompt-sensitive, brief “short” diagnostic prompts underperform relative to “native” report-style prompts~\cite{wald2025comprehensive}. Concurrently, pan-organ CT–VLMs trained at biobank scale have appeared: \emph{Percival} pretrains contrastively on $>400$k CT–report pairs spanning thorax/abdomen/pelvis and yields strong frozen-encoder classifiers (respiratory mean AUROC $\approx 0.73$; cardiovascular $\approx 0.76$ with external cardiomegaly $0.83$ on CT-RATE dataset), with variable but positive gains across abdominal/genitourinary categories~\cite{beeche2025pan}.

Abdomen CT introduces multi-phase protocols and long-tail labels. Recent CT–VLMs (e.g., MERLIN, CT-GLIP) report competitive zero-shot transfer, but supervised and anatomy-aware baselines under fixed protocols remain scarce ~\cite{blankemeier2024merlin, lin2024ct}. ORACLE–CT fills this gap: an encoder-agnostic head that performs organ-restricted softmax pooling with optional Organ-Scalar Fusion, trained end-to-end in a fixed, auditable pipeline and evaluated side-by-side with GAP, global attention, and masked variants introduced next.

 % \emph{Merlin} pretrains on 3D CT with report/EHR supervision and shows competitive zero-shot transfer (F1 $0.74$ internal, $0.65$ external)~\cite{blankemeier2024merlin}. \emph{CT-GLIP} constructs organ-level pairs using multi-organ masks and organwise text, reporting AUROC $0.719$ and F1 $0.526$ on a private 16-class set~\cite{lin2024ct}. 

 % Consistent with clinical prior work, class-wise performance skews higher for \emph{diffuse} vs.\ \emph{focal} abnormalities—lungs: effusion/emphysema AUCs $>0.80$ vs.\ nodules $\sim0.65$; liver: steatosis/biliary dilation $>0.80$ vs.\ lesions $\sim0.73$ and hepatobiliary calcifications $<0.70$; kidneys: stones/atrophy $>0.80$ vs.\ lesions/cysts $\sim0.68$–$0.70$~\cite{tushar2021classification}.

\section{Method: ORACLE\textendash CT}
\label{sec:method}

\begin{figure*}[t]
  \centering
  \includegraphics[width=\linewidth]{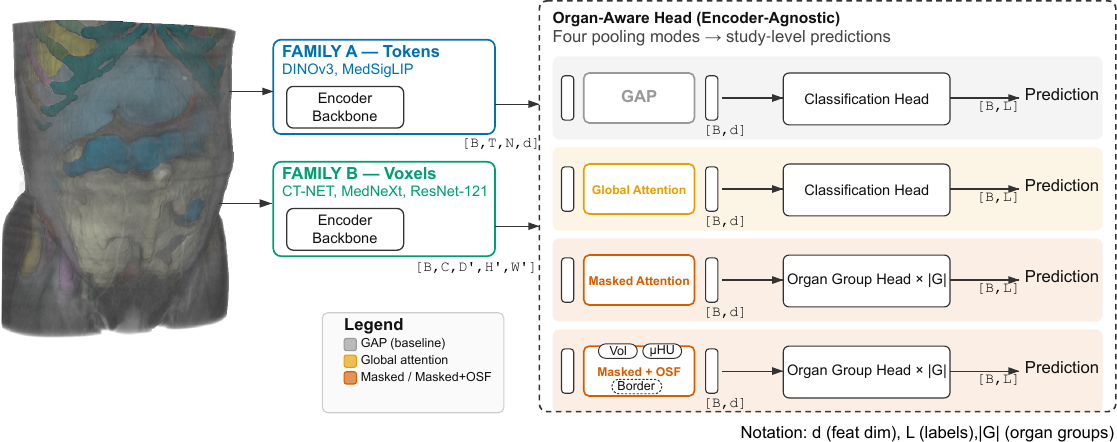}
\caption{\textbf{ORACLE--CT overview.} Two encoder families feed a shared organ-aware head: \emph{Family A} 2.5D token towers (DINOv3, MedSigLIP) and \emph{Family B} native 3D trunks (CT-NET, I3D-ResNet-121, MedNeXt-3D). The head supports four pooling modes: \emph{GAP}, \emph{Global attention} (mask-free unary softmax over the lattice), \emph{Masked attention} (softmax restricted to organ masks; one head per organ group plus an \texttt{other} head), and \emph{Masked + OSF} (append organ volume/HU/border scalars). Outputs are study-level predictions. Organ merges and label groups are in Appendix~\Cref{tab:a1-ctrate-merges,tab:a2-merlin-merges}.}
  \label{fig:overview}
\end{figure*}

\subsection{Problem Setup}
\label{sec:problem}

A CT study is a 3D volume \(X\in\mathbb{R}^{D\times H\times W}\) (axial slices \(D\), height \(H\), width \(W\)). We consider a \textbf{multi-label} setting with \(L\) study-level disease labels. Ground-truth targets are \(y\in\{0,1,-1\}^{L}\), where \(-1\) denotes missing/unknown labels, we predict calibrated label probabilities  $\hat{\mathbf p}\in[0,1]^L$. An encoder \(E_{\theta}\) extracts local features from \(X\), and an aggregation head \(A_{\phi}\) maps them to logits \(z\in\mathbb{R}^{L}\):
\[
z \;=\; A_{\phi}\!\big(E_{\theta}(X)\big), \qquad \hat{\mathbf p}\;=\;\mathrm{sigmoid}(z)\in[0,1]^L.
\]

% When stated, we apply post-hoc calibration (temperature scaling) by replacing \(z\) with \(z/T\) before the sigmoid.

\subsection{Encoders as Feature Adapters}
\label{sec:backbones}

We treat the encoder as a \emph{feature adapter} that maps a CT volume to a spatial lattice of local features. Let \(\Omega\) be the index set of the encoder’s final feature lattice and let \(\{u_i\}_{i\in\Omega}\) with \(u_i\in\mathbb{R}^d\) denote the local features consumed by our aggregation head. Our contribution lies in the aggregation schemes; the head operates only on \(\{u_i\}\) and is therefore \textbf{encoder-agnostic}—any backbone that yields a feature lattice can be plugged in.

\noindent We consider two common substrates (others are compatible): (i) a \emph{token lattice} produced by 2.5D transformers (after dropping CLS/register tokens), and (ii) a \emph{voxel lattice} produced by native 3D networks as shown in Table~\ref{tab:backbone-taxonomy}. We write \(i\in\Omega\) for a lattice location (token or voxel) and flatten its coordinates for brevity.

\begin{table}[t]
\centering
\small
\setlength{\tabcolsep}{6pt}
\renewcommand{\arraystretch}{1.02}
\caption{Illustrative backbones used in our experiments. The aggregation head is identical across all.}
\label{tab:backbone-taxonomy}
\begin{tabular}{lc}
\toprule
Backbone (example) & Feature substrate \\
\midrule
DINOv3~\cite{simeoni2025dinov3} & Token lattice (2D/2.5D) \\
MedSigLIP~\cite{zhai2023sigmoid} & Token lattice (2D/2.5D) \\
I3D--ResNet-121~\cite{he2016deep,carreira2017quo} & Voxel lattice (3D) \\
MedNeXt-3D~\cite{roy2023mednext,liu2022convnet} & Voxel lattice (3D) \\
CT-Net~\cite{draelos2021machine} & Voxel lattice (3D) \\
\bottomrule
\end{tabular}
\end{table}

\subsection{From Feature Lattices to Aggregation Heads}
\label{sec:agg-overview}

Given an encoder that yields a lattice of local features \(\{u_i\}_{i\in\Omega}\), we compare four \emph{encoder-agnostic} heads in a staged ladder to isolate content weighting and anatomy priors (Fig.~\ref{fig:overview}): (i) \textbf{GAP}—uniform averaging with a linear classifier (Eq.~\ref{eq:gap-head}); (ii) \textbf{global attention(mask-free, unary)}—a scalar score per location with softmax over \(\Omega\) (Eqs.~\ref{eq:global-attn-weights}--\ref{eq:global-attn-head}); (iii) \textbf{organ-masked pooling}—the same operator but softmax normalized \emph{within} each organ’s support, followed by per-organ classification (Eqs.~\ref{eq:masked-alpha}--\ref{eq:masked-pool-classify}); and (iv) \textbf{organ-masked + OSF}—augmenting the pooled organ feature with compact mask-derived scalars (volume, mean HU, truncation) before classification (Eq.~\ref{eq:osf-simple}).

\subsubsection{Mask-free Heads}
\label{sec:heads-maskfree}

\paragraph{Global Average Pooling (GAP).}
Given \(\{u_i\in\mathbb{R}^d\}_{i\in\Omega}\), we form a uniform average and classify:
\begin{equation}
  h \;=\; \frac{1}{|\Omega|}\sum_{i\in\Omega} u_i \;\in\; \mathbb{R}^{d},
  \qquad
  z \;=\; W^{\mathrm{global}} h + b^{\mathrm{global}} \;\in\; \mathbb{R}^{L}.
  \label{eq:gap-head}
\end{equation}
Here \(W^{\mathrm{global}}\!\in\!\mathbb{R}^{L\times d}\) and \(b^{\mathrm{global}}\!\in\!\mathbb{R}^{L}\).
GAP is \emph{feature-only}, scales linearly in \(|\Omega|\), is invariant to lattice density,
and serves as our supervised baseline for the \textbf{multi-label} setting.

\paragraph{Global Attention (mask-free, content-weighted).}
To move beyond uniform pooling, we assign a learned weight to each location using a
\emph{unary} scorer (no Q/K/V), then softmax over the whole lattice:
\begin{equation}
  \alpha_i \;=\; s_{\mathrm{glob}}(u_i), \qquad
  w_i \;=\; \frac{\exp(\alpha_i/\tau)}{\sum_{j\in\Omega}\exp(\alpha_j/\tau)}, \quad \tau>0,
  \label{eq:global-attn-weights}
\end{equation}
This is classic \emph{soft attention pooling} over the full lattice (unary scoring + softmax over $\Omega$) as in  ~\cite{bahdanau2014neural,luong2015effective,xu2015show}, and unlike Transformer style Q--K--V feature mixing~\cite{vaswani2017attention,wang2018non}.

\begin{equation}
  h \;=\; \sum_{i\in\Omega} w_i\,u_i \in \mathbb{R}^{d},
  \qquad
  z \;=\; W^{\mathrm{global}} h + b^{\mathrm{global}} \in \mathbb{R}^{L}.
  \label{eq:global-attn-head}
\end{equation}
Here \(s_{\mathrm{glob}}:\mathbb{R}^d\!\to\!\mathbb{R}\) is a tiny per-location map
(e.g., Linear\((d{\to}1)\) for tokens or \(1{\times}1{\times}1\) Conv3D for voxels),
and the weights satisfy \(w_i\!\ge\!0\) and \(\sum_{i}w_i\!=\!1\).
Compared to GAP, which averages uniformly, global attention \emph{learns} a content-weighted
mixture that can emphasize salient regions while remaining mask-free and encoder-agnostic.
The head adds only \(\mathcal{O}(|\Omega|)\) compute and reduces to GAP when
all \(\alpha_i\) are equal (or as \(\tau\!\to\!\infty\)).

\subsubsection{Organ-aware Heads}
\label{sec:heads-organ}

Mask-free pooling (GAP/Global Attention) can overweight off-organ cues. We therefore add an \emph{anatomy prior} that localizes evidence to clinically meaningful regions while keeping the encoder generic. Concretely, raw segmentation classes \(\mathcal{C}\) are merged into a small set of \emph{organ groups} \(\mathcal{O}\) (\eg., all lung lobes \(\rightarrow\) \texttt{lungs}; left/right kidneys \(\rightarrow\) \texttt{kidneys}; \texttt{stomach}\(\vee\)\texttt{esophagus} \(\rightarrow\) \texttt{stomach\_esophagus}). For each raw class \(c \in \mathcal{C}\), let \(S_c \in \{0,1\}^{D\times H\times W}\) denote its binary segmentation mask on the native CT grid (1 inside class \(c\), 0 otherwise). Let \(\pi:\mathcal{C}\to\mathcal{O}\) map classes to groups and define \(M_o=\bigvee_{c:\,\pi(c)=o} S_c\) on the native CT grid. We dilate \(M_o\) by \(r_o\) mm (metric space) to absorb boundary uncertainty, then resample to the encoder lattice \(\Omega\) (tokens or voxels) to obtain indicators \(m_{o,i}\in\{0,1\}\) and support \(\Omega_o=\{\,i\in\Omega:\, m_{o,i}=1\,\}\).

Each study-level label \(\ell\in\mathcal{L}\) is \emph{anchored} to exactly one group via a fixed map \(\kappa:\mathcal{L}\to\mathcal{O}\cup\{\texttt{other}\}\), yielding disjoint label sets \(\{\mathcal{L}_o\}_{o\in\mathcal{O}}\) and \(\mathcal{L}_{\texttt{other}}\). Unlike mask-free heads (single global classifier for all \(L\) labels), the organ-aware design uses a \emph{bank of heads}: one per group \(o\) for \(\mathcal{L}_o\), plus an \texttt{other} head for labels without a stable segmentation region. Organ groups and dilation radii $r_o$ (mm) are defined in Appendix~\Cref{tab:a1-ctrate-merges,tab:a2-merlin-merges}. This procedure is encoder-agnostic; only mask-to-lattice alignment differs across backbones.

\textbf{Organ-masked attention.}
Given \(\{u_i\}_{i\in\Omega}\) and masks \(m_{o,i}\), each organ group \(o\) applies a lightweight
\emph{per-group unary scorer} \(s_o:\mathbb{R}^d\!\to\!\mathbb{R}\) (Linear\((d{\to}1)\) on tokens or
\(1{\times}1{\times}1\) Conv3D on voxels; no Q/K/V, \(\mathcal{O}(|\Omega|)\) cost), a temperature \(\tau_o>0\),
and optional inside/outside priors \((\beta_o^{\mathrm{in}},\beta_o^{\mathrm{out}})\):
\begin{equation}
  \tilde{\alpha}_{o,i} \;=\; s_o(u_i) \;+\; \beta_o^{\mathrm{in}}\,m_{o,i} \;+\; \beta_o^{\mathrm{out}}(1-m_{o,i}).
  \label{eq:masked-alpha}
\end{equation}
Attention is \emph{normalized within} the organ support via a masked softmax,
\begin{equation}
  w_{o,i}
  \;=\;
  \frac{\exp(\tilde{\alpha}_{o,i}/\tau_o)\, m_{o,i}}
       {\sum_{j\in\Omega_o}\exp(\tilde{\alpha}_{o,j}/\tau_o) + \varepsilon},
  \qquad \varepsilon>0,
  \label{eq:masked-softmax}
\end{equation}
followed by per-organ pooling and classification:
\begin{equation}
  h_o \;=\; \sum_{i\in\Omega_o} w_{o,i}\,u_i,
  \qquad
  z_o \;=\; W^{(o)} h_o + b^{(o)} \in \mathbb{R}^{|\mathcal{L}_o|}.
  \label{eq:masked-pool-classify}
\end{equation}
If \(\Omega_o\) is empty at the pooling resolution, we fall back to uniform pooling over \(\Omega\) for that organ.
Labels in \(\mathcal{L}_{\texttt{other}}\) use the mask-free global attention head.

\textbf{Organ--Scalar Fusion (Masked+OSF).}
We extend the organ-masked head by adding a few simple, mask-derived scalars per organ and then
predicting with a small head.

For each organ \(o\), we extract three mask-derived scalars: (i) a \emph{normalized volume} \(v_o\) (z-scored over the training set), (ii) a \emph{mean HU} \(\bar{\mu}^{\mathrm{HU}}_o\) (after standard clipping, then z-scored), and (iii) a binary \emph{border-contact flag} \(b_o\in\{0,1\}\) indicating whether the (dilated) mask touches the scan field-of-view boundary. We collect these as \(u_o=[\,v_o,\,\bar{\mu}^{\mathrm{HU}}_o,\,b_o\,]\) (additional clinically useful scalars can be appended).

From masked attention we already have a per-organ spatial feature \(h_o\in\mathbb{R}^d\).
If the organ is flagged as truncated (\(b_o{=}1\)), we \emph{softly down-weight} \(h_o\) by a single learned
gate (one scalar per organ group) so the model relies a bit less on potentially biased spatial evidence.
We then \emph{append} the scalars to the (possibly down-weighted) feature and predict the organ’s labels with
a small head:
\begin{equation}
\label{eq:osf-simple}
\begin{aligned}
\hat{h}_o \;&=\; [\,h_o;\,u_o\,] \;\in\; \mathbb{R}^{d+k},\\
z_o \;&=\; W^{(o)} \hat{h}_o + b^{(o)} \;\in\; \mathbb{R}^{|\mathcal{L}_o|}.
\end{aligned}
\end{equation}

Here \(W^{(o)}\!\in\!\mathbb{R}^{|\mathcal{L}_o|\times(d+k)}\) and \(b^{(o)}\!\in\!\mathbb{R}^{|\mathcal{L}_o|}\).
When \(k{=}0\) (no scalars), \eqref{eq:osf-simple} reduces exactly to the masked-attention classifier.

\paragraph{Implementation details.}
A complete implementation on the DINOv3 token backbone, covering 3D volume to 2D slices framing, token–lattice formation,
mask projection/alignment, scalar computation and normalization is
provided in Appendix~\Cref{app:dinov3:tokenization}

\subsection{Training Objective and Calibration}

\noindent\textbf{Training loss.}
We minimize a per–label Binary Cross Entropy (BCE) with-logits loss, normalized by the total effective label weight for stable scale. Observed labels contribute with unit weight. Labels marked as missing are ignored for an initial burn-in of \(n_{\text{burn}}\) epochs and are then included as weak negatives with a linearly increasing weight over \(n_{\text{ramp}}\) epochs up to a small cap \(w_{\max}\). This schedule avoids early over-penalizing uncertain supervision and improves stability, calibration, and recall. Class imbalance is handled via conservative positive-class reweighting.

\noindent\textbf{Training \& evaluation.}
All models share the same preprocessing, splits, loss, calibration, and fixed decision-threshold protocol—varying only the aggregation mode (GAP, Global, Masked, Masked{+}OSF) and the encoder family (token vs.\ voxel). Model selection uses validation loss. Per–label temperature scaling and F1–optimal thresholds are fit on the validation set and then frozen for test and external evaluation. Metrics exclude missing targets (\(-1\)) from denominators, following prior work~\cite{blankemeier2024merlin}. We report macro AUROC, AUPRC, F1, and Balanced Accuracy and per class AUROC/AUPRC. Complete hyperparameters appear in Appendix~\Cref{tab:train-shared}.

\section{Experiments and Results}
\label{sec:experiments}

\subsection{Datasets and Splits}
\label{sec:data}

\begin{table*}[t]
\centering
% \small
% \setlength{\tabcolsep}{6pt}
% \renewcommand{\arraystretch}{1.05}
\caption{\textbf{Datasets and I/O geometry.} Organ group counts list only \emph{masked} groups (excluding \texttt{other}); full merges and per-group dilations are in Appendix ~\Cref{tab:a1-ctrate-merges} and ~\Cref{tab:a2-merlin-merges}). All volumes are clipped to $[-1000,1000]$ and normalized to $[0,1]$ (Sec.~\ref{sec:data}).}
\label{tab:data-brief}
\begin{tabular}{lrrrrcc}
\toprule
Dataset & Train & Val & Test & Organ groups (masked) & Resampled Spacing (mm) & Shape \\
\midrule
CT--RATE & 43{,}738 & 3{,}409 & 3{,}040 & 4 (+\texttt{other}) & $1.5{\times}1.5{\times}1.5$ & $224^3$ \\
MERLIN   & 15{,}175 & 5{,}018 & 5{,}082 & 13 (+\texttt{other}) & $3{\times}1.5{\times}1.5$   & $160{\times}224{\times}224$ \\
RAD--ChestCT (ext.) & \multicolumn{2}{c}{---} & 3630 & 4 (+\texttt{other}) & $1.5{\times}1.5{\times}1.5$ & $224^3$ \\
\bottomrule
\end{tabular}
\end{table*}

% \noindent{\textbf{CT\textendash RATE (chest)}
% Large single\textendash institution chest CT corpus with 18 study\textendash level findings; we follow the official label map and patient\textendash disjoint splits. After removing non\textendash chest volumes as in prior work~\cite{lee2025unified}, the final counts are shown in Table~\ref{tab:data-brief}. For organ-aware pooling, we collapse fine-grained segmentation masks into a small set of chest organ groups that define the pooling regions; details are in Appendix ~\Cref{tab:a1-ctrate-merges}.

% \noindent \textbf{{RAD\textendash ChestCT (external)}}
% Public chest CT test set from Duke University Medical Center~\cite{draelos2021machine}. We harmonize its 84 labels to the CT\textendash RATE 16\textendash label subset using the \emph{same mapping protocol as prior work} for fair comparison~\cite{hamamci2024developing}: \eg \textit{Calcification} = $\max(\textit{arterial wall},\,\textit{coronary})$; \textit{Mosaic attenuation} is absent in RAD\textendash ChestCT and thus excluded. External evaluation is strictly transferred: per\textendash label temperature scaling and F1\textendash optimal thresholds are fit on CT\textendash RATE validation and applied unchanged to RAD\textendash ChestCT. Organ\textendash mask merges and metric dilations mirror the CT\textendash RATE configuration. \emph{Per\textendash label prevalence differences across the harmonized 16\textendash label space (CT\textendash RATE vs.\ RAD\textendash ChestCT) are visualized in Appendix\ } \Cref{fig:prevalence_ctrate_test_vs_radchest_test}.

\noindent\textbf{Chest datasets.} \emph{Internal: CT--RATE}—18 study-level findings with official patient-disjoint splits; we follow the released label map, remove non-chest volumes as in prior work~\cite{lee2025unified}, and report final counts in Table~\ref{tab:data-brief}. For organ-aware pooling, fine-grained masks are collapsed into a small set of chest organ groups that define pooling regions (Appendix.~\Cref{tab:a1-ctrate-merges}). \emph{External: RAD--ChestCT}—public chest CT test set~\cite{draelos2021machine} harmonized to the CT--RATE 16-label protocol using the same mapping as prior work~\cite{hamamci2024developing} (e.g., \textit{Calcification} $=\max(\textit{arterial wall},\,\textit{coronary})$; \textit{Mosaic attenuation} absent and excluded). External evaluation is strictly transferred: per-label temperature scaling and F1-optimal thresholds are fit on CT--RATE validation and applied unchanged to RAD--ChestCT; organ-mask merges and dilation radii mirror CT--RATE. 

% Prevalence differences across the harmonized 16 labels are shown in Appendix.~\Cref{fig:prevalence_ctrate_test_vs_radchest_test}.

\noindent \textbf{Abdomen dataset - MERLIN} Large abdomen–focused corpus with 30 study-level labels as  positive, negative or missing ~\cite{blankemeier2024merlin}. We use the released patient-disjoint splits (counts in \Cref{tab:data-brief}). For organ grouping, TotalSegmentator classes~\cite{wasserthal2023totalsegmentator} are merged into \textbf{13} organ groups (with fixed metric-space dilations $r_o$ in mm), plus an additional category \texttt{other}. Here, \texttt{other} collects labels that cannot be assigned to any organ group because (i) the relevant anatomy has no corresponding segmentation class, or (ii) the finding is not well localized (\eg, diffuse/systemic or protocol/device related). The full, fixed mapping of labels to organ groups (including which labels fall under \texttt{other}) is provided in Appendix~\Cref{tab:a2-merlin-merges}. 

% Class prevalence across MERLIN is shown in Appendix~\Cref{fig:merlin_prevalence_stacked_counts}.

\noindent{\textbf{Preprocessing and Augmentations}-} Unless noted, intensities are clipped to $[-1000,1000]$ HU and min–max normalized to $[0,1]$ dataset-specific voxel spacing and volume dimensions are listed in Table~\ref{tab:data-brief}. CT–RATE uses random in-plane rotations and per-axis flips. MERLIN uses small 3D affine transforms (in-plane rotation, translation, scaling), gamma jitter, and low-variance noise, with left–right, anterior–posterior, and cranio–caudal flips disabled to preserve laterality (\eg, spleen left, liver right).
\subsection{Chest CT - Classification Results}

\begin{table*}[t]
\centering
\small
\setlength{\tabcolsep}{6pt}
\renewcommand{\arraystretch}{1.12}
\caption{\textbf{Chest results on CT--RATE (internal) and RAD--ChestCT (external).} Prior VLM baselines are taken \emph{as reported} by ~\cite{wald2025comprehensive}—their strongest \emph{linear–probe} (frozen encoder + linear head) results. In contrast, we evaluate \emph{supervised} encoders (GAP shown) under one fixed, auditable protocol. Per-label temperatures/thresholds are fit once on CT--RATE validation and applied unchanged to CT--RATE test and to RAD--ChestCT, which we harmonize to the CT--RATE 16-label map. For comparability with~\cite{wald2025comprehensive}, AUROC/AUPRC are reported as percentages in this table only.}
\label{tab:chest-main}
\begin{tabular}{lcccccccc}
\toprule
& \multicolumn{4}{c}{\textbf{CT--RATE (internal test)}} & \multicolumn{4}{c}{\textbf{RAD--ChestCT (external)}} \\
\cmidrule(lr){2-5}\cmidrule(lr){6-9}
Method & AUROC $\uparrow$ & AUPRC $\uparrow$ & F1 $\uparrow$ & BA $\uparrow$
       & AUROC $\uparrow$ & AUPRC $\uparrow$ & F1 $\uparrow$ & BA $\uparrow$ \\
\midrule
\multicolumn{9}{l}{\emph{Prior work (VLM, linear--probe references)}}\\
CT--CLIP        & 61.21 & 25.96 & 34.49 & 57.66 & 54.05 & 28.77 & 39.08 & 52.65 \\
CT--FM          & 82.14 & 53.54 & 55.56 & 73.51 & 68.49 & 42.41 & 47.55 & 61.95 \\
Merlin          & 82.62 & 54.81 & 56.69 & 74.28 & 70.91 & 45.30 & 49.35 & 64.34 \\
COLIPRI\text{-}C & 84.15 & 57.41 & 57.99 & 74.99 & 72.66 & 48.86 & 51.31 & 65.17 \\
\midrule
\multicolumn{9}{l}{\emph{Ours (supervised GAP, backbone sweep)}}\\
CT\text{-}Net           & 79.59 & 49.32 & 51.04 & 71.61 & 68.14 & 42.30 & 44.30 & 61.61 \\
MedSigLIP               & 83.61 & 58.17 & 56.24 & 73.93 & 74.44 & 50.40 & 47.76 & 65.80 \\
MedNeXt\text{-}3D       & 84.41 & 58.37 & 57.76 & 75.75 & 72.87 & 48.39 & 48.42 & 65.86 \\
DINOv3                  & 84.52 & 59.17 & 57.24 & 74.30 & 76.25 & 52.59 & 50.11 & 66.33 \\
\textbf{Inflated ResNet\text{-}121} & \textbf{85.74} & \textbf{61.73} & \textbf{59.91} & \textbf{76.12} &
\textbf{76.45} & \textbf{54.32} & \textbf{51.80} & \textbf{68.39} \\
\bottomrule
\end{tabular}
\end{table*}

\begin{table}[t]
\centering
\small
\setlength{\tabcolsep}{6pt}
\renewcommand{\arraystretch}{1.10}
\caption{\textbf{Segmentation-guided chest classification (AUROC).} ORACLE\textendash CT applies organ-masked attention to the best GAP baseline (Inflated ResNet\textendash121). CT\textendash RATE protocol; RAD\textendash ChestCT harmonized to 16 labels.}
\label{tab:chest-seg}
\begin{tabular}{lcc}
\toprule
& \textbf{CT\textendash RATE} & \textbf{RAD\textendash ChestCT} \\
Method & AUROC $\uparrow$ & AUROC $\uparrow$ \\
\midrule
fVLM      & 0.78 & 0.68 \\
SegVL     & 0.81 & 0.72 \\
Uniferum  & 0.83 & \textemdash \\
\midrule
\textbf{Ours (ORACLE\textendash CT)} & \textbf{0.86} & \textbf{0.76} \\
\bottomrule
\end{tabular}
\end{table}

\noindent\textbf{Supervised GAP baselines (SOTA comparison).}
On \emph{CT--RATE}, our supervised GAP setup already achieves \textbf{state-of-the-art} results within our fixed protocol and reported against CT--VLM linear-probe references: \textbf{Inflated ResNet--121 (GAP)} leads with \textbf{AUROC 85.74 / AUPRC 61.73 / F1 59.91 / BA 76.12}, with DINOv3 and MedNeXt close behind.
Under external shift (\emph{RAD--ChestCT}), the same GAP baseline remains strong (\textbf{AUROC 76.45 / F1 51.80}), reinforcing our thesis that calibrated, anatomy-aware supervision—without large-scale Vision Language pretraining—already delivers competitive triage performance. We report full internal/external results and fair VLM linear-probe comparisons in Table~\ref{tab:chest-main}. Per class AUROC/AUPRC scores are presented in Appendix ~\Cref{fig:chest_heatmaps}.

\noindent\textbf{Segmentation–guided upgrade (masked attention).}
Applying our \emph{organ–masked attention} head atop the best supervised backbone (Inflated ResNet–121) yields \(\Delta\)AUROC \(= +0.42\) and \(\Delta\)F1 \(= +0.47\) on \textsc{CT–RATE} vs.\ its \textsc{GAP} baseline, and matches AUROC with a small F1 gain on \textsc{RAD–ChestCT} (Table~\ref{tab:chest-seg}). We run the full four–mode ablation (\textsc{GAP} \(\rightarrow\) Global \(\rightarrow\) Masked \(\rightarrow\) Masked+OSF) on \textsc{MERLIN} (abdomen; 13 organ groups), where fine-grained organ grouping particularly benefits our organ-aware design; chest uses only four groups, yet we still see targeted gains, notably for \emph{hiatal hernia}. Chest results per class AUROC/AUPRC are in Appendix.~\cref{fig:ct-rate-four-modes,fig:radchest-four-modes}.

% \paragraph{Summary.}
% (i) A well–tuned supervised GAP baseline is strong and transfers well on external test set; (ii) organ–aware pooling further improves internal AUROC/F1 and holds parity externally.

\subsection{Abdomen CT Classification Results} 

\noindent\textbf{Supervised vs.\ zero-shot on \emph{Merlin}.}
Table~\ref{tab:merlin_main_comparison} reports macro AUROC/AUPRC/F1/BA on the official \emph{Merlin} test split.
As the released \emph{Merlin}~\cite{blankemeier2024merlin} weights do \emph{not} ship with zero-shot prompts, we constructed class-wise positive and negative phrasings using a large language model, full list in Appendix ~\Cref{tab:merlin_zero_shot_prompts}) and ran zero-shot inference from the authors' checkpoint for a fair comparison on the \emph{full} test set (their reported results are on a class-balanced subset).
Across all five encoders, our fully supervised models surpass the reproduced zero-shot baseline; the DINOv3 encoder is the best overall, absolute AUROC/AUPRC values are present in Appendix ~\Cref{fig:merlin_models_comparison_heatmaps}.

% \noindent\textbf{Four aggregation modes on DINOv3.}
% Figure~\ref{fig:dumbbell_all25_four_modes_AUROC} (25 mask-eligible classes) and Appx.~\Cref{fig:merlin_dinov3_modes_comparison_heatmaps} show that organ conditioning outperforms \textsc{GAP}, and adding scalars further helps. In particular, \emph{Masked+OSF} (mean HU + volume) lifts the \emph{macro} AUROC from \textbf{0.83} to \textbf{0.85} \;(\(+0.02\) absolute; \(\sim\!2.4\%\) relative) under the same splits, preprocessing, and calibration. The improvements concentrate on size/morphology–driven labels (\eg, “–megaly”, obstruction, hydronephrosis) while remaining positive on most mask-eligible classes. Encoder unfreezing adds a further small, consistent boost; we analyze its interaction with scalar type and region choice in Ablations~\Cref{sec:ablations}.

\begin{figure*}[t]
  \centering
  \includegraphics[width=\textwidth]{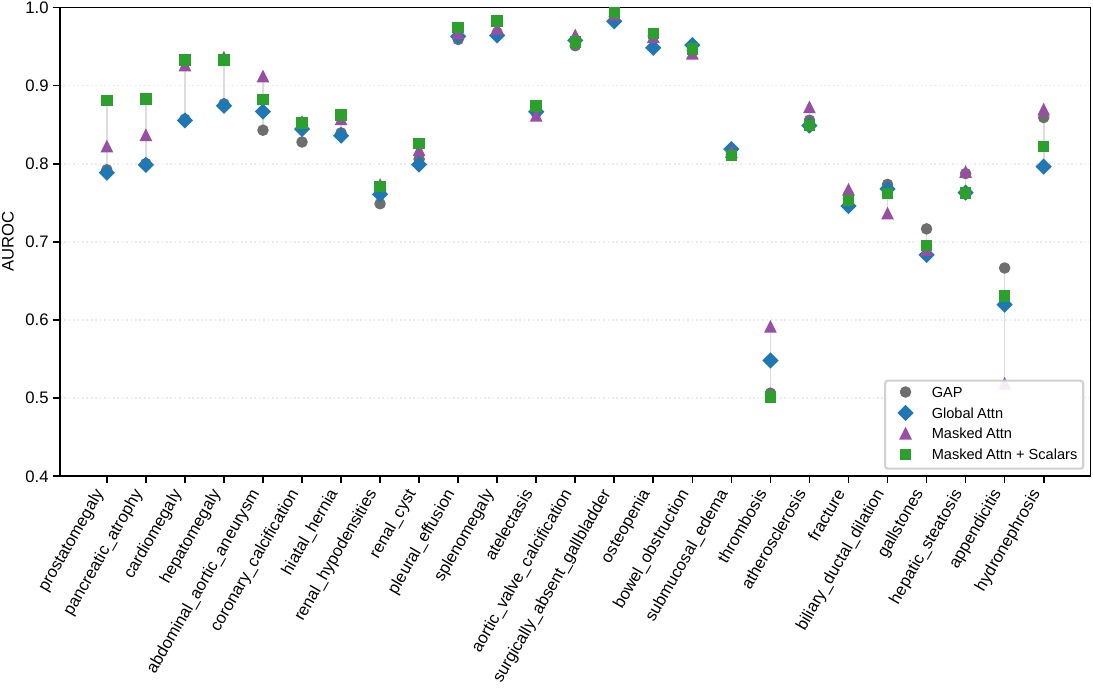}
\caption{\textbf{MERLIN (abdomen), DINOv3: per-class AUROC across four aggregation modes.} We compare \emph{GAP}, \emph{Global attention} (mask-free), \emph{Organ-masked attention}, and \emph{Masked+OSF} (masked attention + organ scalars: mean HU, volume). Shown are the 25 mask-eligible classes (the 5 labels without stable organ masks are omitted). \emph{Classes are sorted by the absolute AUROC gain of Masked+OSF over GAP}. Masked+OSF provides the most consistent improvements—especially for size/morphology–driven findings at the left—while organ-masked attention alone already lifts most organ-tied labels.}
  \label{fig:dumbbell_all25_four_modes_AUROC}
\end{figure*}

\noindent\textbf{Four aggregation modes on DINOv3.}
Table~\ref{tab:merlin_modes} shows both mask-based variants outperform \textsc{GAP} and global attention; \emph{Masked+Scalar} is slightly better than masked attention (AUROC 0.85 vs 0.84; AUPRC 0.68 vs 0.66); with the improvements notably on size driven labels(\eg, “–megaly”). As masked attention depends on segmentation quality, labels lacking reliable masks (\eg, appendicitis, gallbladder pathology) can favor \textsc{GAP}, see class-wise trends in Fig.~\ref{fig:dumbbell_all25_four_modes_AUROC}. To address this, Ablations~\Cref{sec:ablations} also evaluates bounding-box regions in place of masks.

% \noindent\textbf{Summary.} On MERLIN, supervised, anatomy-aware pooling is a reliable upgrade over both GAP and zero-shot transfer; augmenting masked attention with simple, interpretable organ scalars (volume, HU) delivers the largest, most clinically meaningful gains, especially for size-driven conditions.

\begin{table}[t]
\centering
\setlength{\tabcolsep}{6pt}
\small

\caption{\textbf{MERLIN test Results.} Top row: zero-shot baseline on the official MERLIN test set; original prompts are unavailable, so we re-created them (see Appendix~\Cref{tab:merlin_zero_shot_prompts}). Rows below the rule: our supervised GAP baselines across five encoder backbones.}
\label{tab:merlin_main_comparison}
\begin{tabular}{lcccc}
\toprule
\textbf{Method} & \textbf{AUROC} $\uparrow$ & \textbf{AUPRC} $\uparrow$ & \textbf{F1} $\uparrow$ & \textbf{BA} $\uparrow$ \\
\midrule
Merlin Zero-Shot & 0.72 & 0.50 & 0.47 & 0.64 \\
\midrule
\textbf{Ours (CT-Net)}                & 0.75 & 0.51 & 0.54 & 0.66 \\
\textbf{Ours (MedNeXt)}              & 0.78 & 0.54 & 0.55 & 0.66 \\
\textbf{Ours (ResNet-121)}           & 0.81 & 0.59 & 0.59 & 0.69 \\
\textbf{Ours (MedSigLip)}           & 0.82 & 0.61 & 0.59 & 0.70 \\
\textbf{Ours (DINOv3)}               & \textbf{0.83} & \textbf{0.64} & \textbf{0.63} & \textbf{0.72} \\
\bottomrule
\end{tabular}
\label{tab:merlin-main}
\end{table}

\begin{table}[t]
\centering
\setlength{\tabcolsep}{6pt}
\small

\caption{\textbf{MERLIN mode ablations (DINOv3).} Compare GAP, global attn, masked attn, and masked attn+scalar (+volume, HU, border-touch); mask-based modes beat GAP/global, with masked+scalar showing the best results.}

\label{tab:merlin_modes}
\begin{tabular}{lcccc}
\toprule
\textbf{DINOv3 Modes} & \textbf{AUROC} $\uparrow$ & \textbf{AUPRC} $\uparrow$ & \textbf{F1} $\uparrow$ & \textbf{BA} $\uparrow$ \\
\midrule
GAP Baseline            & 0.83 & 0.64 & 0.63 & 0.72 \\
Global Attn        & 0.83 & 0.63 & 0.63 & 0.73 \\
Masked Attn        & 0.84 & 0.66 & 0.63 & 0.73 \\
Masked Attn + OSF & \textbf{0.85} & \textbf{0.68} & \textbf{0.66} & \textbf{0.74} \\
\bottomrule
\end{tabular}
\end{table}

\subsection{Qualitative Evidence}
\label{sec:qual}

\begin{figure*}[t]
  \centering
  % Replace the path below with your actual PDF path
  \includegraphics[width=\textwidth]{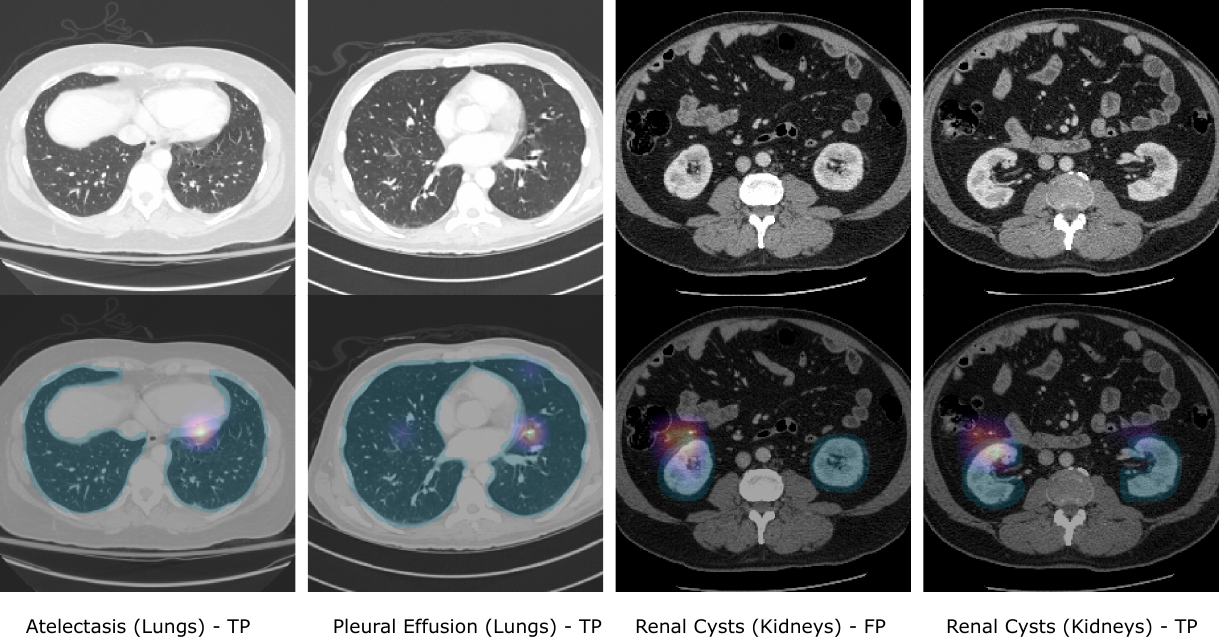}
  \caption{\textbf{Qualitative organ--masked attention on MERLIN.}
  Columns show four study examples: Atelectasis (TP), Pleural effusion (TP), Renal cyst (FP), and Renal cyst (TP).
  Top row: axial CT slice. Bottom row: organ mask (cyan) with masked--attention heatmap (magma).
  Maps depict \emph{organ-level pooling weights} used for study-level classification (not lesion segmentation): they highlight where evidence within the organ contributed most to the logit.
  The FP renal-cyst case illustrates organ-faithful but misleading evidence.}
  \label{fig:qual-attn-merlin}
\end{figure*}

We visualize organ-masked attention for four MERLIN studies using DINOv3 in masked attention mode in Figure \ref{fig:qual-attn-merlin}. Each column shows an axial CT slice (top) and an overlay (bottom) with the organ mask (cyan) and the masked-attention heatmap (magma). For lung findings (atelectasis, pleural effusion), attention remains within the lungs. For renal cysts, the TP example shows kidney-confined evidence; the FP example also shows kidney-confined attention but an incorrect positive prediction. These maps depict organ-level pooling weights used to form the study-level logits—they provide anatomical context for where the model drew evidence, not pixel-accurate lesion segmentations or guarantees of correctness.

\subsection{Ablations}
\label{sec:ablations}

\noindent\textbf{(A) Backbone freezing.}
All \textsc{Merlin} results in Table~\ref{tab:merlin-main} use \emph{unfrozen} backbones. Comparing frozen vs.\ unfrozen backbone for DINOv3 and MedSigLIP, unfreezing consistently lifts AUROC/AUPRC, with larger and more uniform gains for DINOv3. Improvements concentrate on multi-organ, morphology/size–driven findings (\eg, “–megaly”, bowel obstruction, hydronephrosis, surgically absent gallbladder), while highly localized or primarily HU-driven patterns show smaller or mixed changes. Net: unfreezing helps, but for focal targets most of the lift still comes from segmentation-guided attention (masked pooling + scalars). Per-class trends appear in Appendix~\Cref{fig:app-ablation_backbone_freeze}.

\noindent\textbf{(B) Scalar fusion: HU-only vs.\ Volume-only.}
Fixing the region to the full \emph{Segmentation Mask}, we compare HU-only vs.\ organ Volume-only fusion (both against the same GAP baseline). Volume tends to help size/morphology findings (\eg \emph{cardiomegaly}, \emph{hepatomegaly}, \emph{splenomegaly}, \emph{pancreatic atrophy}), while HU favors intensity/density patterns (\eg calcifications, thrombosis). Full per-class heatmaps are in Appendix \Cref{fig:app-ablation-auroc-hu-vs-vol}.

\noindent\textbf{(C) Mask region: Segmentation Mask vs.\ Bounding Box.}
Masked pooling and OSF depend on segmentation; gains scale with region \emph{fidelity} and \emph{footprint}. We therefore compare using the full \emph{Mask} vs.\ a tight \texttt{Bounding Box} as the aggregation region. Both generally beat GAP, but the winner is class-dependent: for \emph{appendicitis}, \texttt{Bounding Box} better captures peri-appendiceal context; for \emph{biliary ductal dilation}, thin boundary-adjacent ducts benefit from periportal context. \emph{Rule of thumb:} use \texttt{Bounding Box} for small/tubular or boundary-adjacent targets; use \emph{Mask} for diffuse within-organ signals; otherwise choose by validation. Per-class heatmaps (AUROC/AUPRC) are in Appendix~\Cref{fig:app-ablation-auprc-mask-vs-bbox}.

\section{Limitations and Future Work}

\textbf{Limitations.} Our approach assumes reliable multi-organ segmentation; errors especially on small/tubular or truncated structures propagate to masked pooling/OSF. Masks are fixed pre-processing (no uncertainty or joint seg–class training). OSF uses uniform fusion of z-volume and mean HU, which can dilute focal signal. Attention maps aid auditability but are not guaranteed faithful. We did not fully tune backbones/schedules.

\textbf{Future work.}
(i) \emph{Disease-aware, gated scalars:} replace mean HU with robust summaries and enable sparse per-label fusion gates.
(ii) \emph{Faithfulness/robustness:} deletion–insertion curves, counterfactual mask edits, and calibration under masking.
(iii) \emph{Mask reliability \& fallbacks:} estimate mask quality and marginalize/switch across mask, padded box, or expanded regions.
(iv) \emph{Tuning \& scaling:} systematic hyper-parameter sweeps, with parameter-efficient adapters.

\section{Conclusion}
We presented ORACLE-CT, a unified, supervised benchmark for CT triage that fixes data splits, label maps, and training protocol while varying two axes—aggregation (GAP, Global Attention, Masked, Masked+OSF) and backbone family (2.5D tokens vs.\ native 3D voxels). Across chest and abdomen, organ-masked pooling improves organ-tied labels and calibration, with the masked+OSF head adding small, consistent gains for size-driven conditions. 
We will release code, configs, and checkpoints to enable comparisons and extensions. We hope this provides a practical baseline and a clear recipe for deploying calibrated, anatomy-aware CT triage systems.

%% file: sec/X_suppl.tex
\clearpage
\setcounter{page}{1}

\maketitlesupplementary

\begin{figure*}[t]
  \centering
  \includegraphics[width=\textwidth]{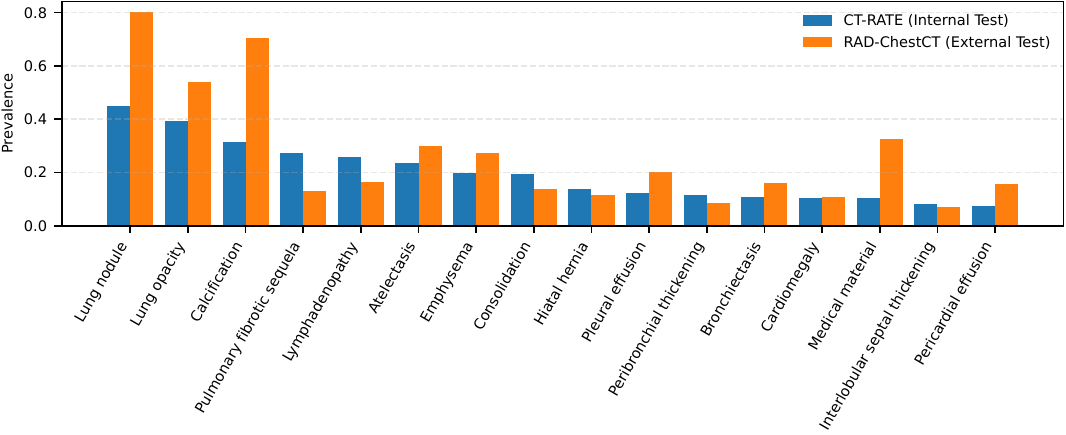}
  \caption{Prevalence comparison on the \textbf{harmonized 16-label chest space}. Bars show per-label prevalence in \textbf{CT--RATE (Internal Test)} and \textbf{RAD--ChestCT (External Test)}, ordered by CT--RATE prevalence. For visualization, CT--RATE \emph{Calcification} is computed as the union of \emph{Arterial wall calcification} and \emph{Coronary artery wall calcification}, and \emph{Mosaic attenuation pattern} is excluded because it is not annotated in RAD--ChestCT.}
  \label{fig:prevalence_ctrate_test_vs_radchest_test}
\end{figure*}

\begin{figure*}[t]
  \centering
  \includegraphics[width=\textwidth]{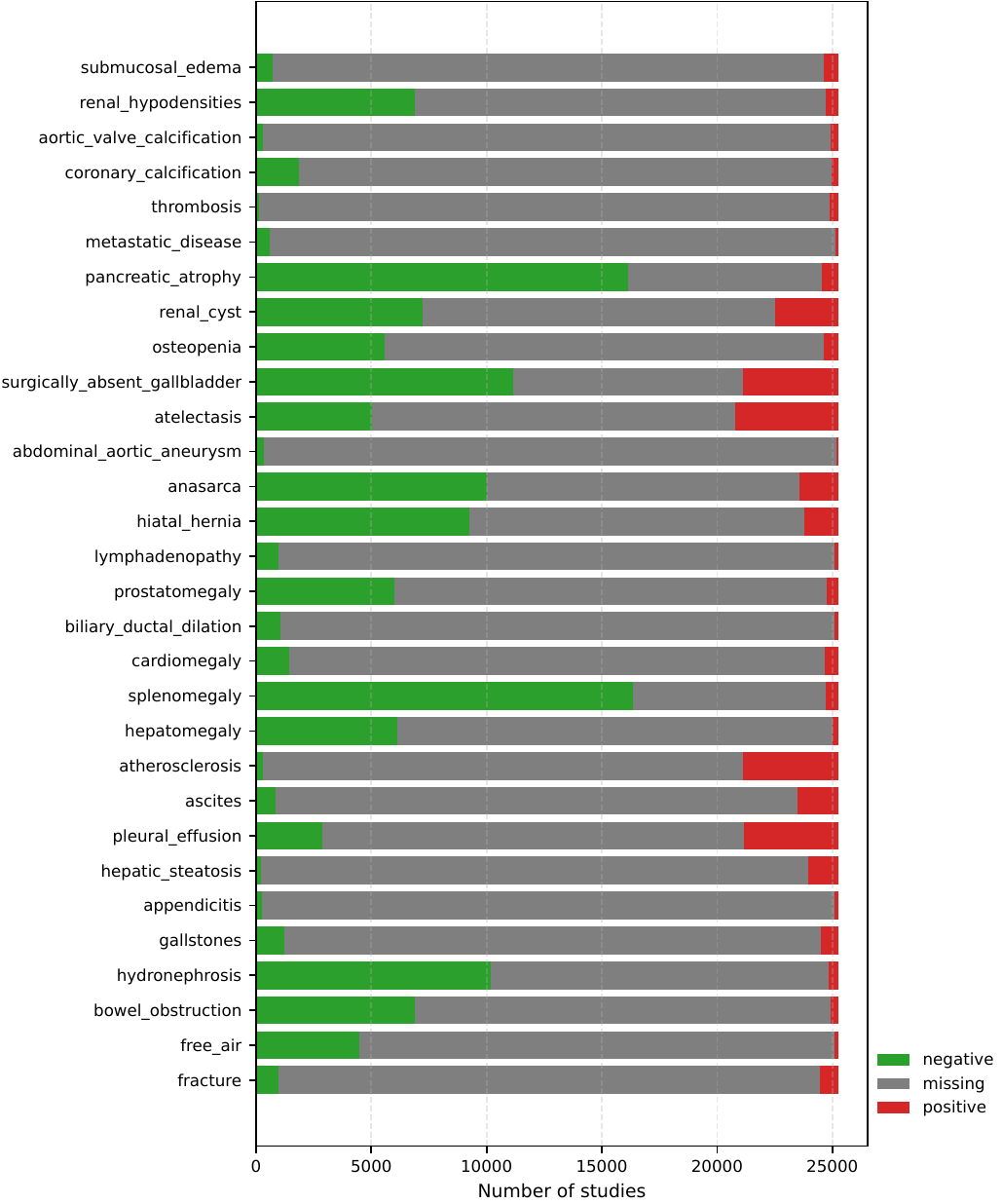}
  \caption{\textbf{MERLIN (abdomen) class prevalence.}
  Horizontal \emph{stacked} bar plot showing, for each of the 30 findings (y--axis), the total number of studies (x--axis) partitioned into \textcolor{green!60!black}{\textbf{negatives (0)}}, \textcolor{red!70!black}{\textbf{positives (1)}}, and \textcolor{gray!60!black}{\textbf{uncertain/missing ($-1$)}}. Counts are aggregated over MERLIN \emph{train+val+test}.}
  \label{fig:merlin_prevalence_stacked_counts}
  \label{app:data-end} % ensure page range A: start--end resolves to the *last* page of Appendix A
\end{figure*}

\begin{figure*}[t]
  \centering
  \includegraphics[width=\textwidth]{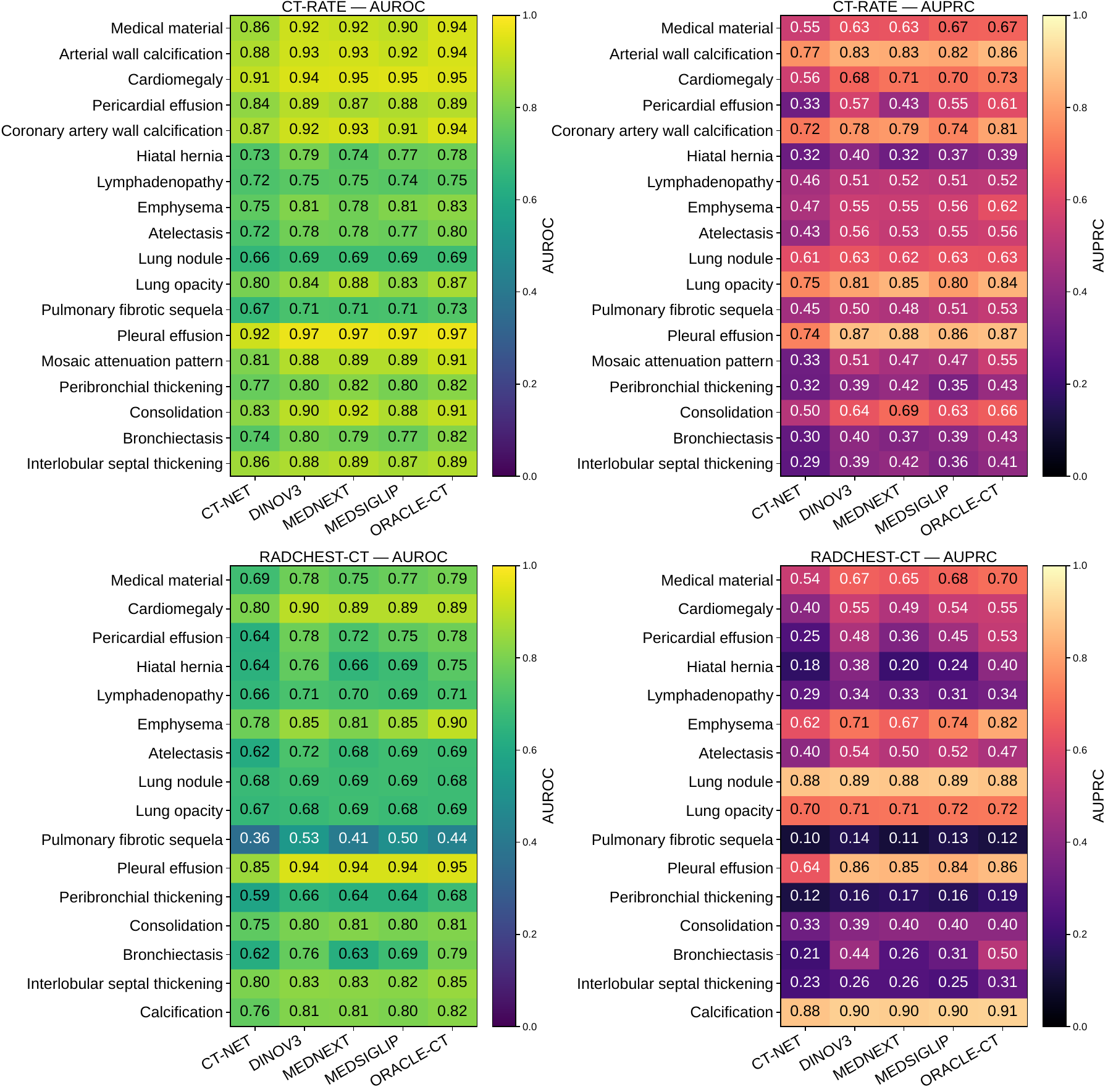}
  \caption{\textbf{Per-class AUROC and AUPRC heatmaps on CT-RATE (top) and RADChest-CT (bottom).} Left column shows AUROC; right column shows AUPRC. Rows are disease labels and columns are models (\eg, \mbox{CT-NET}, \mbox{DINOV3}, \mbox{MEDNEXT}, \mbox{MEDSIGLIP}, \mbox{ORACLE-CT}). Each cell prints the exact score, while color encodes magnitude using distinct, metric-specific colormaps (see colorbars).}
  \label{fig:chest_heatmaps}
\end{figure*}

\begin{figure*}[t]
  \centering
  \includegraphics[width=\textwidth]{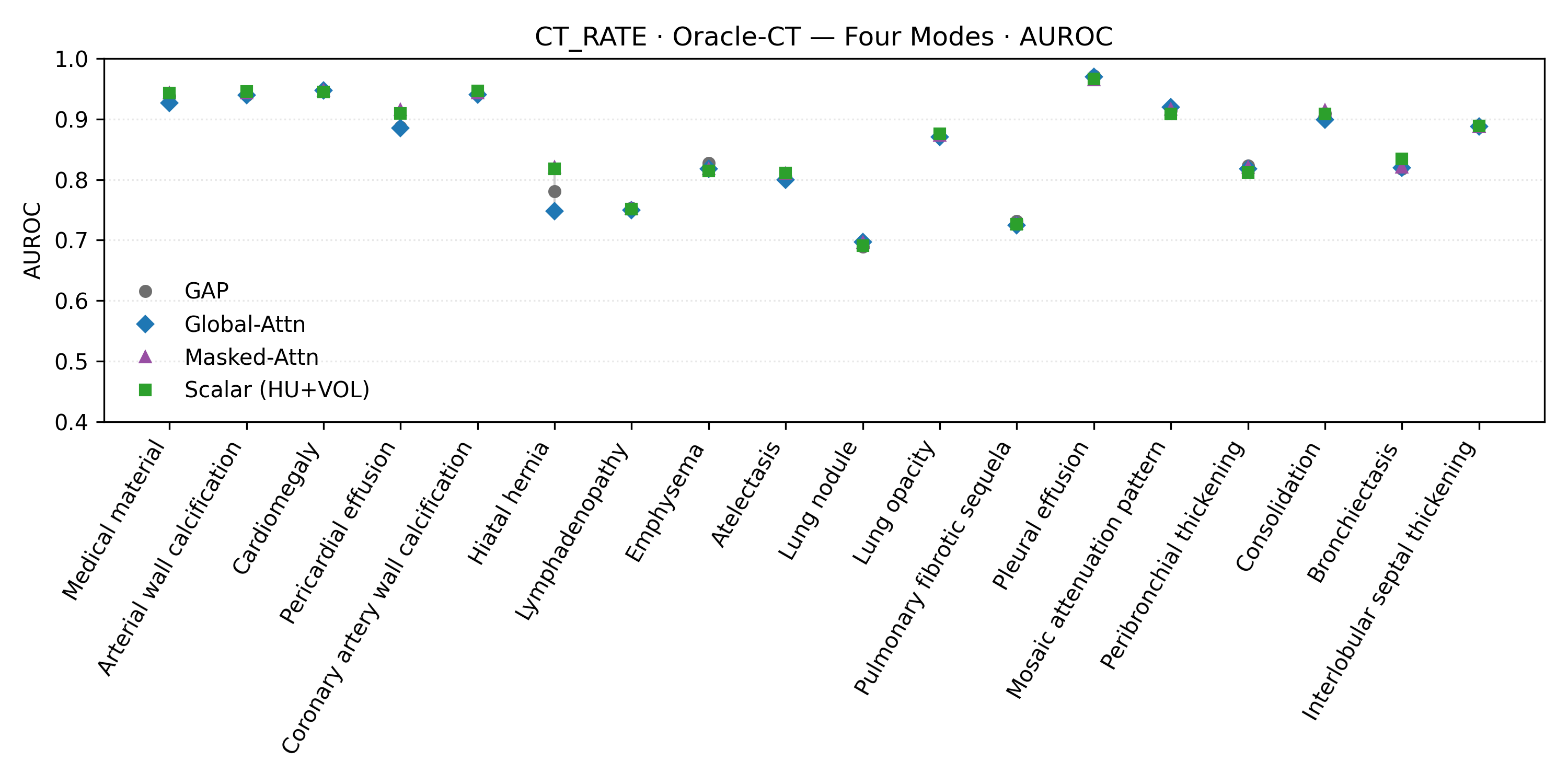}
  \caption{\textbf{CT–RATE, ORACLE-CT mode ablation (per-label).}
  Four modes (GAP, Global, Masked, Masked{+}Scalar) show tightly clustered performance; median differences are small, reflecting coarse chest anatomy.}
  \label{fig:ct-rate-four-modes}
\end{figure*}

\begin{figure*}[t]
  \centering
  \includegraphics[width=\textwidth]{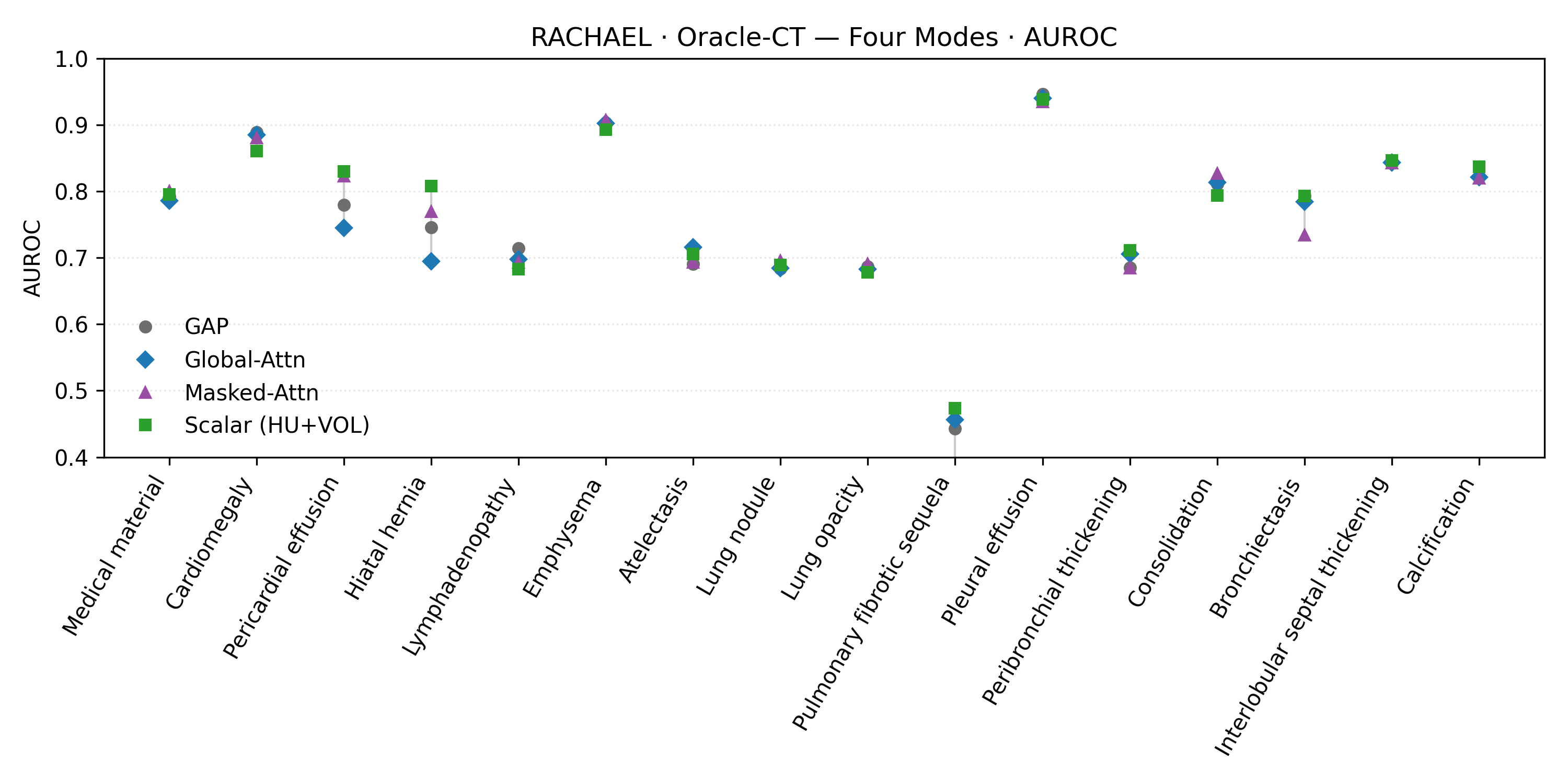}
  \caption{\textbf{RAD-ChestCT, ORACLE-CT mode ablation (per-label).}
  Four modes (GAP, Global, Masked, Masked{+}Scalar) show tightly clustered performance; median differences are small, reflecting coarse chest anatomy.}
   \label{fig:radchest-four-modes}
\end{figure*}

\begin{figure*}[t]
  \centering
\includegraphics[width=\textwidth,height=0.90\textheight,keepaspectratio]{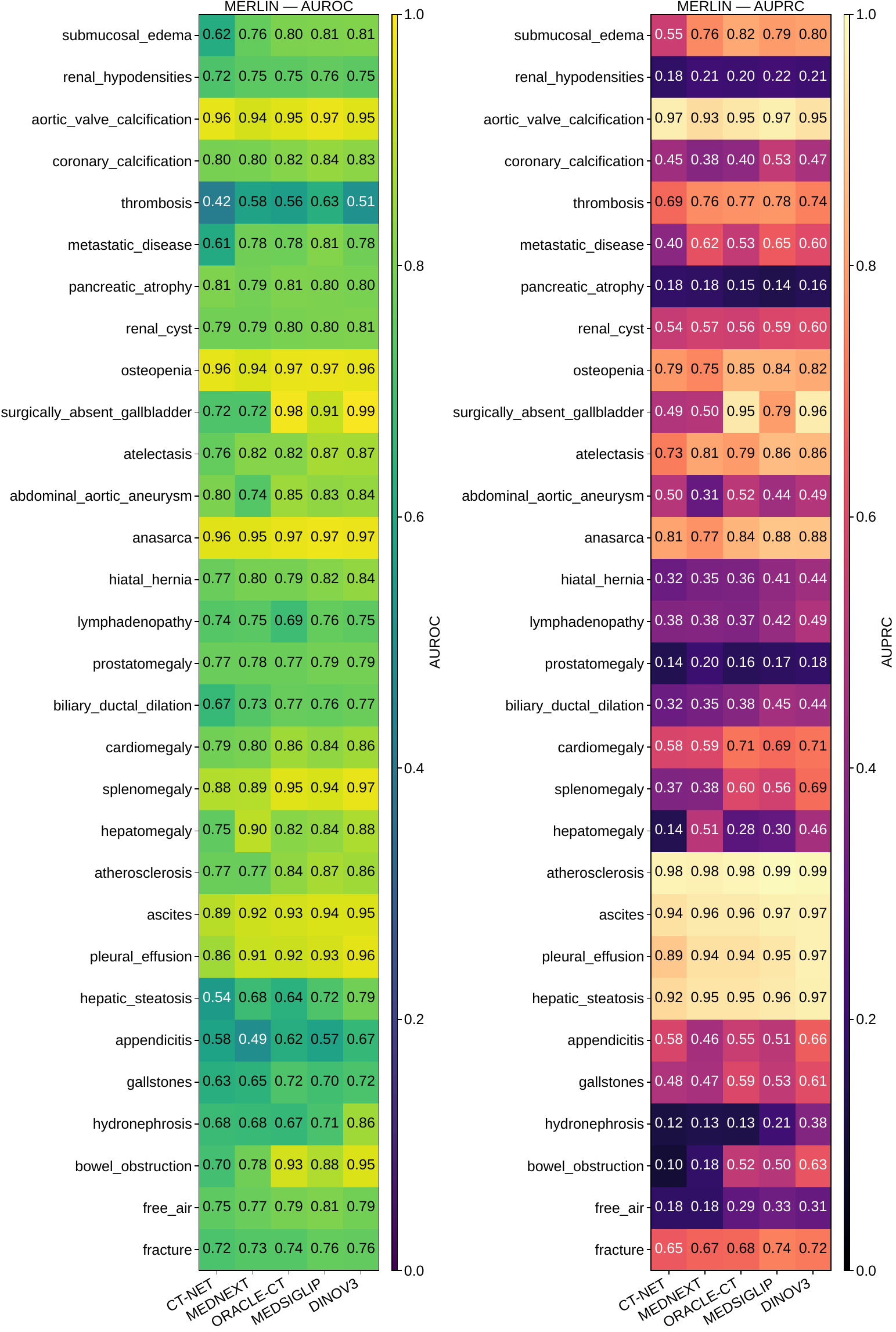}
  \caption{\textbf{MERLIN test set — GAP baseline (Global Average Pooling).} Per-class heatmaps compare five encoders with unfrozen backbones. \emph{Left:} AUROC. \emph{Right:} AUPRC. Columns (left→right): \textbf{CT-NET}, \textbf{MEDNEXT}, \textbf{ORACLE-CT}, \textbf{MEDSIGLIP}, \textbf{DINOV3}; rows are the 30 abdominal findings. Overall, \textbf{DINOV3} and \textbf{MEDSIGLIP} lead across most classes, \textbf{ORACLE-CT} is competitive mid-pack, and \textbf{MEDNEXT}/\textbf{CT-NET} trail. High-signal targets (e.g., calcifications, pleural effusion, hepatic steatosis) approach saturation across models, while gaps are larger on lower-prevalence/acute findings (\eg, appendicitis, bowel obstruction, free air, gallstones).}

  \label{fig:merlin_models_comparison_heatmaps}
\end{figure*}

\begin{figure*}[t]
  \centering
  \includegraphics[width=\textwidth,height=0.90\textheight,keepaspectratio]{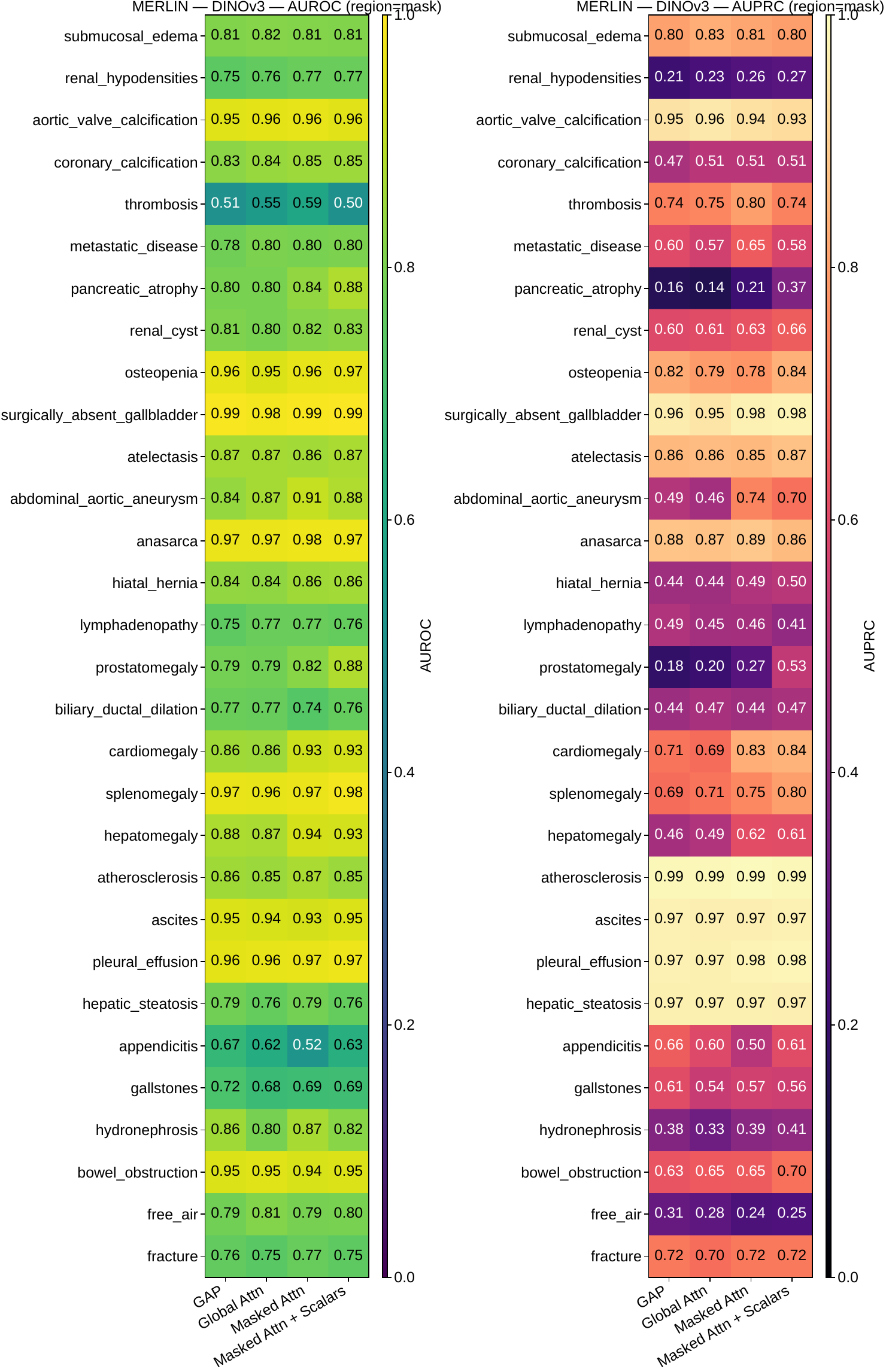}
  \caption{\textbf{DINOv3—Four aggregation modes with organ masks (AUROC/AUPRC, 30 MERLIN classes).} \emph{GAP} (global average pooling baseline), \emph{global\_attn} (unconstrained token attention), \emph{masked\_attn} (attention restricted to the target organ using \texttt{mask\_region=Mask}), and \emph{masked\_attn\_scalar} (masked attention with scalar fusion of organ \emph{volume} and mean \emph{HU}). Columns enumerate the four modes; rows list the 30 abdominal labels. Brighter cells indicate higher AUROC. Organ-aware heads (masked\_attn / masked\_attn\_scalar) confine evidence to anatomy and typically improve class-specific discrimination over GAP, with masked\_attn\_scalar adding small gains by injecting simple volumetric and intensity cues.}
  \label{fig:merlin_dinov3_modes_comparison_heatmaps}
\end{figure*}

% Compact full-width (two-column spanning) Table of Contents
\twocolumn[{
\vspace{0.4em}
{\large\bfseries Table of Contents\par}
\vspace{0.25em}
\begingroup\small

\noindent
\hyperref[app:data-start]{A.\quad Data \& Preprocessing}
\dotfill \pageref{tab:a1-ctrate-merges}\\[-0.15em]

\noindent\hspace*{1.2em}
\hyperref[tab:a1-ctrate-merges]{A.1\; CT--RATE organ merges (table)}
\dotfill \pageref{tab:a1-ctrate-merges}\\[-0.15em]

\noindent\hspace*{1.2em}
\hyperref[tab:a2-merlin-merges]{A.2\; MERLIN organ merges (table)}
\dotfill \pageref{tab:a2-merlin-merges}\\[-0.15em]

\noindent\hspace*{1.2em}
\hyperref[fig:prevalence_ctrate_test_vs_radchest_test]{A.3\; CT--RATE vs RAD--ChestCT prevalence (fig)}
\dotfill \pageref{fig:prevalence_ctrate_test_vs_radchest_test}\\[-0.15em]

\noindent\hspace*{1.2em}
\hyperref[fig:merlin_prevalence_stacked_counts]{A.4\; MERLIN class prevalence (fig)}
\dotfill \pageref{fig:merlin_prevalence_stacked_counts}\\[0.3em]

\noindent
\hyperref[app:results-start]{B.\quad Results}
\dotfill \pageref{fig:chest_heatmaps}\\[-0.15em]

\noindent\hspace*{1.2em}
\hyperref[sec:results-chest]{B.1\; Chest}
\dotfill \pageref{fig:chest_heatmaps}\\[-0.15em]

\noindent\hspace*{2.4em}
\hyperref[fig:chest_heatmaps]{B.1.1\; Chest Model Heatmaps}
\dotfill \pageref{fig:chest_heatmaps}\\[-0.15em]

\noindent\hspace*{2.4em}
\hyperref[fig:ct-rate-four-modes]{B.1.2\; CT-RATE (four modes)}
\dotfill \pageref{fig:ct-rate-four-modes}\\[-0.15em]

\noindent\hspace*{2.4em}
\hyperref[fig:radchest-four-modes]{B.1.3\; RAD--ChestCT (four modes)}
\dotfill \pageref{fig:radchest-four-modes}\\[-0.15em]

\noindent\hspace*{1.2em}
\hyperref[sec:results-abdomen]{B.2\; Abdomen}
\dotfill \pageref{fig:merlin_models_comparison_heatmaps}\\[-0.15em]

\noindent\hspace*{2.4em}
\hyperref[fig:merlin_models_comparison_heatmaps]{B.2.1\; Merlin Models Comparison AUROC/AUPRC [Scalar=Mask]}
\dotfill \pageref{fig:merlin_models_comparison_heatmaps}\\[-0.15em]

\noindent\hspace*{2.4em}
\hyperref[fig:merlin_dinov3_modes_comparison_heatmaps]{B.2.2\; Merlin Model - DINOv3 Mode Comparison AUROC/AUPRC [Scalar=Mask]}
\dotfill \pageref{fig:merlin_dinov3_modes_comparison_heatmaps}\\[0.3em]

\noindent
\hyperref[app:ablation-start]{C.\quad Ablations}
\dotfill \pageref{fig:app-ablation_backbone_freeze}\\[-0.15em]

\noindent\hspace*{1.2em}
\hyperref[sec:ablation-freeze]{C.1\; Merlin Backbone: frozen vs.\ unfrozen (GAP)}
\dotfill \pageref{fig:app-ablation_backbone_freeze}\\[-0.15em]

\noindent\hspace*{1.2em}
\hyperref[sec:ablation-scalar]{C.2\; Merlin Scalars (Mask fixed): HU vs.\ Vol}
\dotfill \pageref{fig:app-ablation-auroc-hu-vs-vol}\\[-0.15em]

\noindent\hspace*{1.2em}
\hyperref[sec:ablation-region]{C.3\; Merlin Region: Mask vs.\ BBox}
\dotfill \pageref{fig:app-ablation-auprc-mask-vs-bbox}\\[-0.15em]

\noindent
\hyperref[app:train-details-start]{D.\quad Training \& Evaluation Details}
\dotfill \pageref{tab:train-shared}\\[-0.15em]

\noindent\hspace*{1.2em}
\hyperref[tab:train-shared]{D.1\; Shared training setup (table)}
\dotfill \pageref{tab:train-shared}\\[-0.15em]

\noindent\hspace*{1.2em}
\hyperref[tab:augmentations]{D.2\; Augmentation presets (table)}
\dotfill \pageref{tab:augmentations}\\[-0.15em]

\noindent\hspace*{1.2em}
\hyperref[tab:model-deltas]{D.3\; Model-specific deltas (table)}
\dotfill \pageref{tab:model-deltas}\\[-0.15em]

\noindent\hspace*{1.2em}
\hyperref[tab:merlin_zero_shot_prompts]{D.4\; Zero-shot prompts — MERLIN (table)}
\dotfill \pageref{tab:merlin_zero_shot_prompts}\\

\noindent
\hyperref[app:dinov3:tokenization]{E.\quad ORACLE-CT with Dinov3 Implementation Notes}
\dotfill \pageref{app:dinov3:tokenization}\\[-0.15em]

\endgroup
}]

\phantomsection\label{app:data-start}

% ---------------- CT-RATE merges table ----------------

\begin{table*}[t]
\centering
\small
\setlength{\tabcolsep}{6pt}
\renewcommand{\arraystretch}{1.08}
\caption{\textbf{CT--RATE (chest):} TotalSegmentator classes $\rightarrow$ merged organ groups $\rightarrow$ dilation (mm) $\rightarrow$ mapped disease labels.}
\label{tab:a1-ctrate-merges}
\begin{tabular}{p{0.30\linewidth} p{0.16\linewidth} p{0.10\linewidth} p{0.40\linewidth}}
\toprule
\textbf{TotalSegmentator classes} & \textbf{Merged organ group} & \textbf{Dilation (mm)} & \textbf{Mapped disease labels} \\
\midrule
\begin{tabular}[t]{@{}l@{}}
\texttt{lung\_upper\_lobe\_left}, \\
\texttt{lung\_lower\_lobe\_left},\\
\texttt{lung\_upper\_lobe\_right},\\
\texttt{lung\_middle\_lobe\_right},\\
\texttt{lung\_lower\_lobe\_right}
\end{tabular}
& \textbf{lung} & 2.0 &
\begin{tabular}[t]{@{}l@{}}
Emphysema, Atelectasis, Lung nodule,\\
Lung opacity, Pulmonary fibrotic sequela, \\
Pleural effusion, Mosaic attenuation pattern,\\
Peribronchial thickening, Consolidation, \\
Bronchiectasis, Interlobular septal thickening
\end{tabular} \\
\cmidrule(lr){1-4}
\texttt{heart}, \texttt{atrial\_appendage\_left}
& \textbf{heart} & 2.0 &
Cardiomegaly, Pericardial effusion, \\
Coronary artery wall calcification \\
\cmidrule(lr){1-4}
\texttt{stomach}, \texttt{esophagus}
& \textbf{stomach\_esophagus} & 1.0 &
Hiatal hernia \\
\cmidrule(lr){1-4}
\texttt{aorta}
& \textbf{aorta} & 2.0 &
Arterial wall calcification \\
\cmidrule(lr){1-4}
\textemdash
& \textbf{other} & --- &
Medical material, Lymphadenopathy \\
\bottomrule
\end{tabular}
\caption*{\emph{Note:} \textbf{other} conditions are not mapped to any organ mask (no consistent organ-localized mask available) and are predicted by a global head without organ masking. Dilation values are metric-space dilations applied to masks prior to pooling.}
\end{table*}

% ---------------- MERLIN merges table ----------------
\begin{table*}[t]
\centering
\small
\setlength{\tabcolsep}{6pt}
\renewcommand{\arraystretch}{1.08}
\caption{\textbf{MERLIN (abdomen):} TotalSegmentator classes $\rightarrow$ merged organ groups $\rightarrow$ dilation (mm) $\rightarrow$ mapped disease labels.}
\label{tab:a2-merlin-merges}
\begin{tabular}{p{0.28\linewidth} p{0.16\linewidth} p{0.10\linewidth} p{0.40\linewidth}}
\toprule
\textbf{TotalSegmentator classes} & \textbf{Merged organ group} & \textbf{Dilation (mm)} & \textbf{Mapped disease labels} \\
\midrule
\texttt{liver} & \textbf{liver} & 3.0 & Hepatic steatosis, Hepatomegaly, Biliary ductal dilation \\
\cmidrule(lr){1-4}
\texttt{gallbladder} & \textbf{gallbladder} & 4.0 & Gallstones, Surgically absent gallbladder \\
\cmidrule(lr){1-4}
\texttt{pancreas} & \textbf{pancreas} & 2.0 & Pancreatic atrophy \\
\cmidrule(lr){1-4}
\texttt{spleen} & \textbf{spleen} & 3.0 & Splenomegaly \\
\cmidrule(lr){1-4}
\texttt{kidney\_left}, \texttt{kidney\_right} & \textbf{kidneys} & 5.0 & Hydronephrosis, Renal cyst, Renal hypodensities \\
\cmidrule(lr){1-4}
\texttt{prostate} & \textbf{prostate} & 4.0 & Prostatomegaly \\
\cmidrule(lr){1-4}
\texttt{stomach}, \texttt{esophagus} & \textbf{stomach\_esophagus} & 3.0 & Hiatal hernia \\
\cmidrule(lr){1-4}
\texttt{small\_bowel}, \texttt{duodenum}, \texttt{colon} & \textbf{bowel} & 3.0 & Appendicitis, Bowel obstruction, Submucosal edema \\
\cmidrule(lr){1-4}
\begin{tabular}[t]{@{}l@{}}
\texttt{lung\_upper\_lobe\_left}, \\
\texttt{lung\_lower\_lobe\_left},\\
\texttt{lung\_upper\_lobe\_right}, \\
\texttt{lung\_middle\_lobe\_right},\\
\texttt{lung\_lower\_lobe\_right}
\end{tabular}
& \textbf{lungs} & 3.0 & Atelectasis, Pleural effusion \\
\cmidrule(lr){1-4}
\texttt{heart}, \texttt{atrial\_appendage\_left} & \textbf{heart} & 4.0 & Cardiomegaly, Coronary calcification, Aortic valve calcification \\
\cmidrule(lr){1-4}
\begin{tabular}[t]{@{}l@{}}
\texttt{aorta}, \texttt{brachiocephalic\_trunk},\\
\texttt{subclavian\_artery L/R},\\
\texttt{common\_carotid\_artery L/R},\\
\texttt{iliac\_artery L/R}
\end{tabular}
& \textbf{arteries} & 3.0 & Abdominal aortic aneurysm, Atherosclerosis \\
\cmidrule(lr){1-4}
\begin{tabular}[t]{@{}l@{}}
\texttt{inferior\_vena\_cava},\\
\texttt{superior\_vena\_cava},\\
\texttt{portal\_vein\_and\_splenic\_vein},\\
\texttt{iliac\_vein L/R},\\
\texttt{brachiocephalic\_vein L/R},\\
\texttt{pulmonary\_vein}
\end{tabular}
& \textbf{veins} & 3.0 & Thrombosis \\
\cmidrule(lr){1-4}
\begin{tabular}[t]{@{}l@{}}
\texttt{Vertebrae C1--S1},\\
\texttt{ribs L/R 1--12},\\
\texttt{sacrum}, \texttt{sternum},\\
\texttt{clavicles L/R}, \texttt{scapula L/R},\\
\texttt{hips L/R}, \texttt{femur L/R},\\
\texttt{humerus L/R}, \texttt{skull}
\end{tabular}
& \textbf{bones} & 0.5 & Osteopenia, Fracture \\
\cmidrule(lr){1-4}
\textemdash & \textbf{other} & --- & Anasarca, Ascites, Free air, Lymphadenopathy, Metastatic disease \\
\bottomrule
\end{tabular}
\caption*{\emph{Note:} \textbf{other} conditions are not mapped to any organ mask (no consistent organ-localized mask available) and are predicted by a global head without organ masking. Dilation values are metric-space dilations applied to masks prior to pooling.}
\end{table*}

% ===================== A.2 Prevalence (no visible subsection header) =====================

% ---- CT-RATE vs RAD-ChestCT prevalence figure ----

% ---- MERLIN prevalence figure (LAST item in Appendix A) ----

%Chest CT_RATE 4 Modes Results on CT_RATE AND RADCHEST 
\phantomsection\label{app:results-start}
\phantomsection\label{sec:results-start}
\phantomsection\label{sec:results-chest}

\label{sec:results-abdomen}

%Ablation
\phantomsection\label{app:ablation-start}
\label{sec:ablation-freeze}
\begin{figure*}[t]
  \centering
\includegraphics[width=\textwidth,height=0.90\textheight,keepaspectratio]{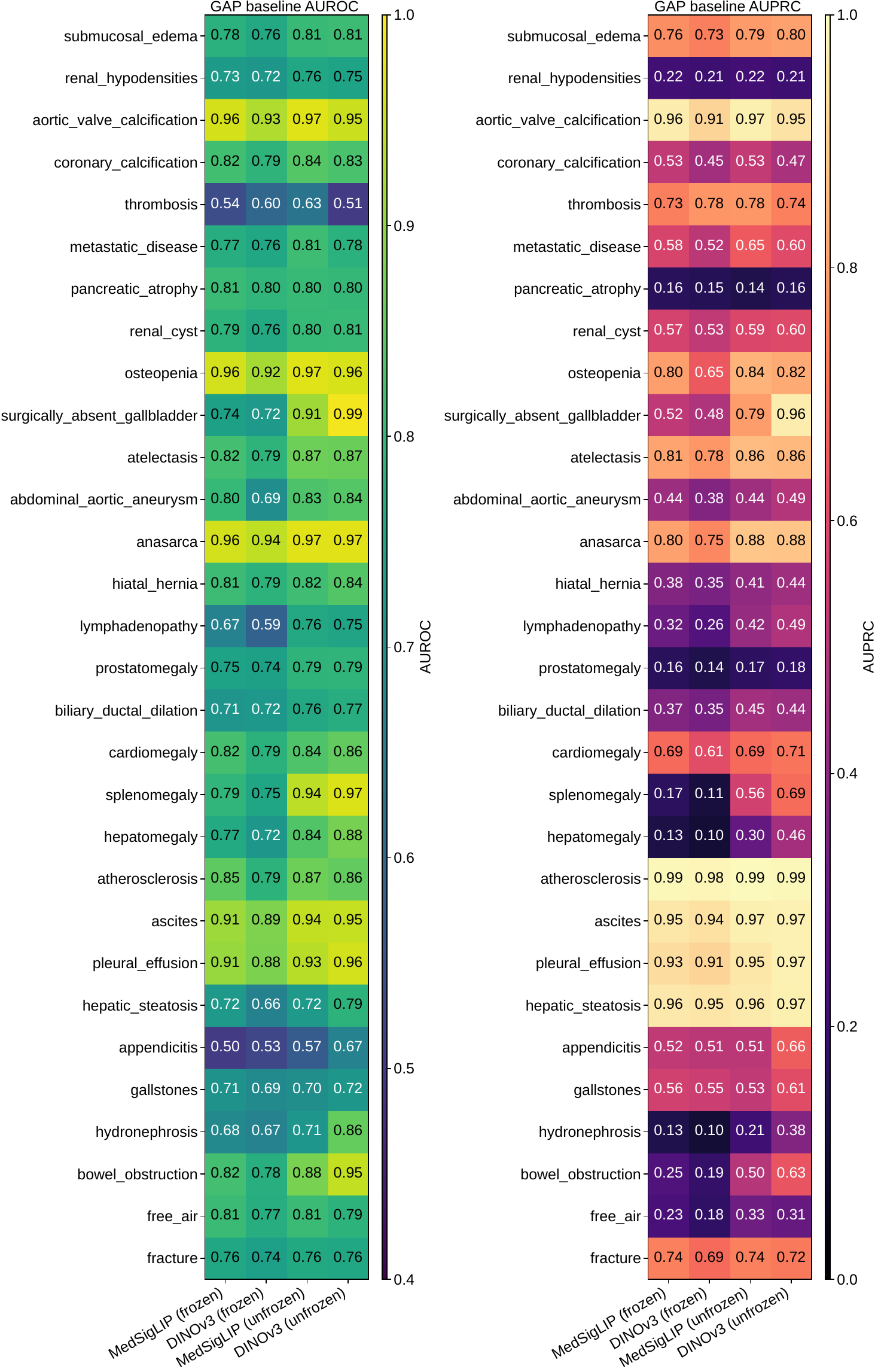}
\caption{\textbf{Backbone freeze vs.\ unfreeze (AUROC/AUPRC; GAP baseline; 30 MERLIN classes).} Absolute per-class AUROC/AUPRC heatmaps for two encoders—DINOv3 and MedSigLIP—trained with a \emph{GAP} head under two regimes: \emph{frozen} backbone and \emph{unfrozen} backbone. Columns enumerate (encoder × freeze state); rows list the 30 abdominal labels. Unfreezing generally raises AUPRC, with the clearest gains on multi-organ and morphology/size–driven findings (e.g., spleno-/hepato-/cardiomegaly, bowel obstruction, surgically absent gallbladder). Highly localized or primarily HU-driven patterns (e.g., pancreatic atrophy, free air) change less, motivating the mask-region and scalar-fusion ablations that follow. Main results elsewhere use unfrozen backbones.}
  \label{fig:app-ablation_backbone_freeze}
\end{figure*}

%%%Ablation Scalar hu vs vol 
\label{sec:ablation-scalar}
\begin{figure*}[t]
  \centering
\includegraphics[width=\textwidth,height=0.90\textheight,keepaspectratio]{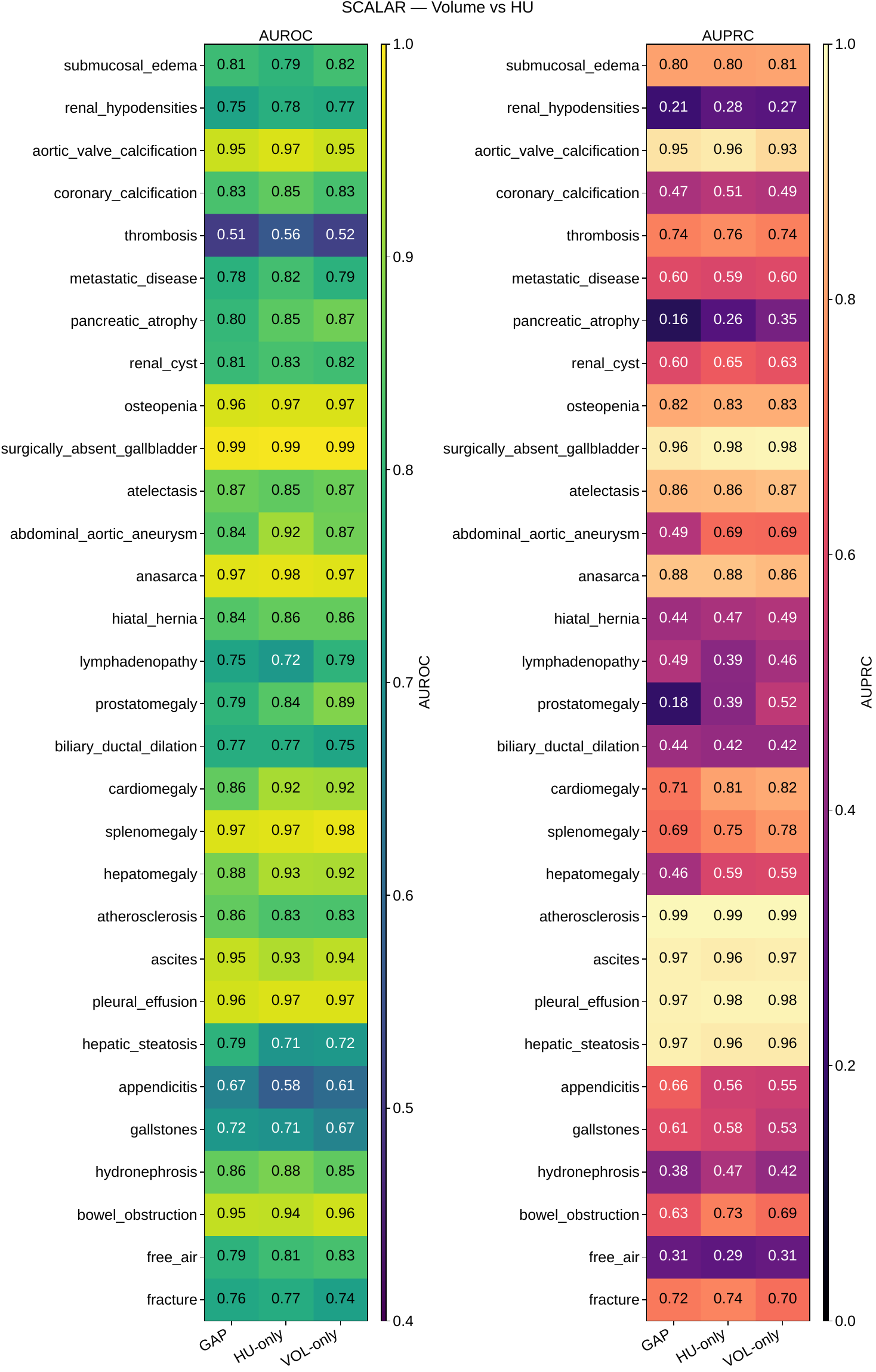}
  \caption{\textbf{Scalar fusion ablation (absolute AUROC; mask region = \texttt{Mask}).} Heatmap of per-class \emph{absolute} AUROC/AUPRC under the \emph{masked\_attn\_scalar} head with the region fixed to the full Segmentation \texttt{Mask}. Columns enumerate scalar variants—\emph{HU-only} vs.\ \emph{Volume-only}—trained with the same encoder and recipe; rows list the 30 MERLIN labels. Volume tends to aid morphology/size–driven findings \eg cardio-/hepato-/splenomegaly, pancreatic atrophy), while HU favors intensity/density patterns (e.g., calcifications, thrombosis). Both variants typically outperform the GAP baseline. Brighter cells indicate higher AUROC.}
  \label{fig:app-ablation-auroc-hu-vs-vol}
\end{figure*}

\label{sec:ablation-region}
\begin{figure*}[t]
  \centering
\includegraphics[width=\textwidth,height=0.90\textheight,keepaspectratio]{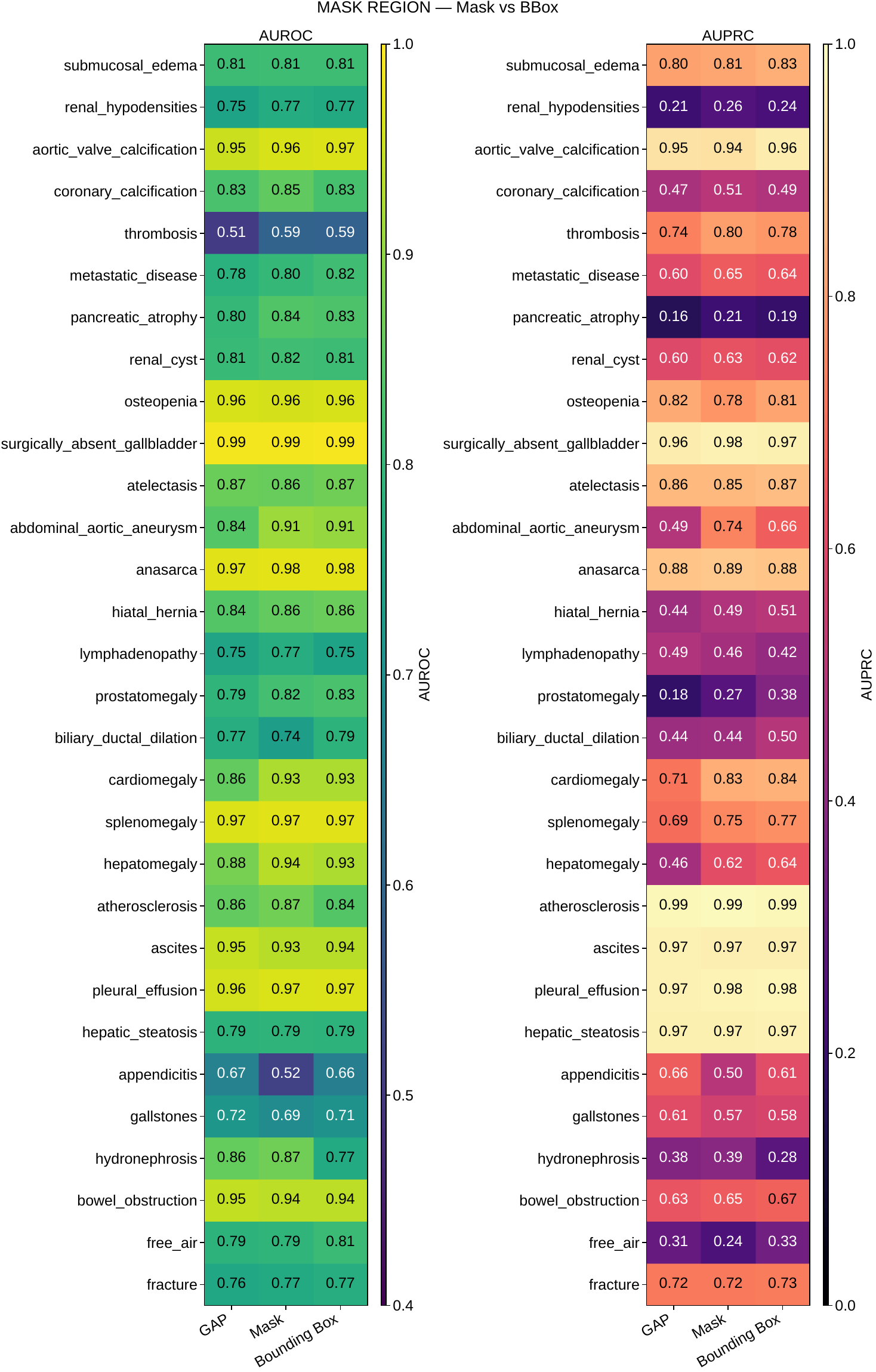}
 \caption{\textbf{Mask region ablation - Mask vs Bounding Box (BBox).}
Heatmap of per-class \emph{absolute} AUROC when the aggregation region for the organ-aware attention head is either the full Segmentation \texttt{Mask} or its tight \texttt{Bounding Box}. Columns enumerate region choice compared to GAP Baseline; rows list the 30 MERLIN labels. Both regions generally outperform the GAP baseline. The preferred region is class-dependent: e.g., \emph{appendicitis} benefits from \texttt{BBox}, which better captures peri-appendiceal context (+26.92\% AUROC; 0.66 vs.\ 0.52), and \emph{biliary ductal dilation} gains from \texttt{BBox} that includes periportal/pancreatic-head context (+6.75\%; 0.79 vs.\ 0.74). Brighter cells indicate higher AUROC.}

  \label{fig:app-ablation-auprc-mask-vs-bbox}
\end{figure*}

\phantomsection\label{app:train-details-start}
% -------- B.1 Shared training setup --------
\begin{table*}[t]
\centering
\footnotesize
\setlength{\tabcolsep}{6pt}
\renewcommand{\arraystretch}{1.06}
\caption{\textbf{Shared training configuration (all models).} Single run with fixed seed; dataset-specific augmentations are summarized in App.~A. Items not listed are identical across datasets and encoders.}
\label{tab:train-shared}
\begin{tabular}{@{}p{0.23\textwidth} p{0.73\textwidth}@{}}
\toprule
\textbf{Item} & \textbf{Setting} \\
\midrule
Seed & \textbf{25} \\
Optimizer & AdamW (\texttt{lr}=0.001, \texttt{weight\_decay}=0.05) \\
LR schedule & Cosine decay with 5-epoch warmup; step-based updates; minimum LR $=0.1\times$ base \\
Parameter groups & Head LR scale $=3.0$; “alpha” params LR scale $=0.3$; group LRs enabled \\
Epochs / Early stop & Max 30 epochs; early stopping on \emph{validation loss}, patience $=10$ \\
Model selection & Best \emph{validation loss} checkpoint (\texttt{checkpoint\_policy: best\_only}) \\
Loss & BCE-uncertain: ignore first 10 epochs; ramp 10 epochs; final uncertain weight $=0.3$ (target $0$) \\
Class weighting & Positive-class weighting enabled (\texttt{use\_pos\_weight: true}); clip at $10.0$ \\
Precision & Mixed precision (AMP) \\
Regularization & Global-norm grad clip $=1.0$; \texttt{attn\_entropy\_reg\_weight}$=0$; \texttt{attn\_coverage\_reg\_weight}$=0$ \\
Parallelism & Distributed Data Parallel (\texttt{ddp: true}); \texttt{cudnn\_benchmark: true} \\
Logging & Log every 1000 steps \\
Transforms & Image/mask transforms applied jointly; masks resampled with nearest; border padding \\
Empty-mask fallback & Uniform pooling over $\Omega$  \\
Calibration & Per-label temperature scaling on validation (\texttt{max\_iter}$=200$, \texttt{min\_count}$=64$) \\
Thresholds & Validation F1-optima, transferred unchanged to test and external sets \\
Augmentations (by dataset) & CT--RATE: \texttt{legacy\_v1} ($k{\times}90^\circ$ rotations; axis flips). \quad MERLIN: \texttt{anatomy\_safe\_v2} (small in-plane affine, mild gamma \& noise; flips \emph{disabled} to preserve laterality). \\
\bottomrule
\end{tabular}
\vspace{-2mm}
\end{table*}

\begin{table*}[t]
\centering
\small
\setlength{\tabcolsep}{6pt}
\renewcommand{\arraystretch}{1.08}
\caption{Augmentation presets and parameters (applied to images and masks together).}
\label{tab:augmentations}
\begin{tabular}{p{0.18\linewidth} p{0.20\linewidth} p{0.58\linewidth}}
\toprule
\textbf{Dataset} & \textbf{Preset} & \textbf{Parameters} \\
\midrule
CT--RATE & \texttt{legacy\_v1} &
\begin{tabular}[t]{@{}l@{}}
\texttt{allow\_rot90: true}, \\
\texttt{allow\_flip\_w: true}, \\
\texttt{allow\_flip\_h: true}, \\
\texttt{allow\_flip\_d: true}
\end{tabular} \\
\cmidrule(lr){1-3}
MERLIN & \texttt{anatomy\_safe\_v2} &
\begin{tabular}[t]{@{}l@{}}
\texttt{p\_affine: 0.35}, \ \texttt{rot\_deg: 10.0}, \ \texttt{translate\_xy: 5.0}, \\
\texttt{scale\_min: 0.95}, \ \texttt{scale\_max: 1.05}, \\
\texttt{allow\_flip\_w: false}, \ \texttt{allow\_flip\_h: false}, \ \\ \texttt{allow\_flip\_d: false}, \\
\texttt{p\_gamma: 0.30}, \ \texttt{gamma\_min: 0.90}, \ \texttt{gamma\_max: 1.10}, \\
\texttt{p\_noise: 0.30}, \ \texttt{noise\_std: 0.01}
\end{tabular} \\
\bottomrule
\end{tabular}
\end{table*}

\begin{table*}[t]
\centering
\small
\setlength{\tabcolsep}{4pt} % tighter padding to avoid overflow
\renewcommand{\arraystretch}{1.05}
\caption{Model–specific settings not shared across all encoders.}
\label{tab:model-deltas}
\begin{tabular}{@{}p{0.24\textwidth} p{0.16\textwidth} p{0.16\textwidth} p{0.18\textwidth} p{0.22\textwidth}@{}}
\toprule
\textbf{Model} & \textbf{Input size} & \textbf{Normalization} & \textbf{Freeze policy} & \textbf{Notes} \\
\midrule
DINOv3 (tokens) & $S\!\in\!\{224,224\}$ & ImageNet & trainable & TRI stride $s{=}1$ \\
MedSigLIP (tokens) & $S{=}224$ & CLIP & trainable & CLIP fallback norm \\
I3D--ResNet--121 (voxels) & $224^3$ & [0-1] & trainable & stage 3/4 pooled \\
MedNeXt--3D (voxels) & $160{\times}224{\times}224$ & [0-1] & trainable & optional pyramid \\
CT--Net (voxels) & $160{\times}224{\times}224$ & [0-1] & trainable & light 3D neck \\
\bottomrule
\end{tabular}
\end{table*}

\label{sec:merlin_zero_shot_prompts}
\begin{table*}[t]
\centering
\scriptsize
\setlength{\tabcolsep}{4pt}
\renewcommand{\arraystretch}{1.1}
\caption{\textbf{Zero-shot prompts used for \emph{Merlin}}. We used these prompts with the released \emph{Merlin} weights to reproduce zero-shot results on the \emph{full} test split. Prompts were generated once with an LLM and fixed before evaluation.}
\label{tab:merlin_zero_shot_prompts}
\begin{tabular}{@{}p{0.24\textwidth} p{0.38\textwidth} p{0.38\textwidth}@{}}
\toprule
\textbf{Class} & \textbf{Positive prompt(s)} & \textbf{Negative prompt(s)} \\
\midrule
\texttt{submucosal\_edema} & submucosal edema; bowel wall edema present & no submucosal edema; no bowel wall edema \\
\texttt{renal\_hypodensities} & renal hypodensity; hypodense renal lesion & no renal hypodensity; no hypodense renal lesion \\
\texttt{aortic\_valve\_calcification} & aortic valve calcification present; calcified aortic valve & no aortic valve calcification \\
\texttt{coronary\_calcification} & coronary artery calcifications; coronary calcification present & no coronary artery calcification; no coronary calcification \\
\texttt{thrombosis} & thrombosis; venous thrombosis; occlusive thrombus; nonocclusive thrombus & no thrombosis; no venous thrombus \\
\texttt{metastatic\_disease} & metastatic disease; metastases present & no metastatic disease; no metastases \\
\texttt{pancreatic\_atrophy} & pancreatic atrophy; atrophic pancreas & no pancreatic atrophy \\
\texttt{renal\_cyst} & renal cyst; simple renal cyst & no renal cyst \\
\texttt{osteopenia} & osteopenia; low bone density & no osteopenia; normal bone mineralization \\
\texttt{surgically\_absent\_gallbladder} & surgically absent gallbladder; post cholecystectomy & gallbladder present; no prior cholecystectomy \\
\texttt{atelectasis} & atelectasis; subsegmental atelectasis & no atelectasis \\
\texttt{abdominal\_aortic\_aneurysm} & abdominal aortic aneurysm; infrarenal AAA & no abdominal aortic aneurysm \\
\texttt{anasarca} & anasarca; diffuse body wall edema & no anasarca; no diffuse body wall edema \\
\texttt{hiatal\_hernia} & hiatal hernia & no hiatal hernia \\
\texttt{lymphadenopathy} & lymphadenopathy; enlarged lymph nodes & no lymphadenopathy; no enlarged lymph nodes \\
\texttt{prostatomegaly} & prostatomegaly; enlarged prostate & no prostatomegaly; prostate normal in size \\
\texttt{biliary\_ductal\_dilation} & biliary ductal dilatation; dilated bile ducts; intrahepatic biliary dilatation & no biliary ductal dilatation; bile ducts normal caliber \\
\texttt{cardiomegaly} & cardiomegaly; enlarged heart & no cardiomegaly \\
\texttt{splenomegaly} & splenomegaly; enlarged spleen & no splenomegaly \\
\texttt{hepatomegaly} & hepatomegaly; enlarged liver & no hepatomegaly \\
\texttt{atherosclerosis} & atherosclerosis; atherosclerotic calcifications & no atherosclerosis; no vascular calcification \\
\texttt{ascites} & ascites; free intraperitoneal fluid & no ascites; no free intraperitoneal fluid \\
\texttt{pleural\_effusion} & pleural effusion; left pleural effusion; right pleural effusion & no pleural effusion \\
\texttt{hepatic\_steatosis} & hepatic steatosis; fatty liver & no hepatic steatosis; no fatty liver \\
\texttt{appendicitis} & appendicitis; inflamed appendix & no appendicitis; normal appendix \\
\texttt{gallstones} & gallstones; cholelithiasis & no gallstones; no cholelithiasis \\
\texttt{hydronephrosis} & hydronephrosis; pelvicaliectasis & no hydronephrosis; no pelvicaliectasis \\
\texttt{bowel\_obstruction} & bowel obstruction; small bowel obstruction & no bowel obstruction \\
\texttt{free\_air} & free air; pneumoperitoneum & no free air; no pneumoperitoneum \\
\texttt{fracture} & fracture; vertebral fracture; rib fracture & no fracture; no acute fracture \\
\bottomrule
\end{tabular}
\end{table*}

\phantomsection\label{app:train-details-end}

\clearpage

\subsection{ORACLE--CT with DINOv3 backbone}
\label{app:dinov3:tokenization}

We detail ORACLE--CT on a 2.5D DINOv3 token tower (\emph{Family A} in Sec.~\ref{sec:backbones}), while preserving the encoder-agnostic aggregation interface in Sec.~\ref{sec:agg-overview}.

\paragraph{TRI construction.}
Given a CT study $X \in \mathbb{R}^{D\times H\times W}$ (axial index first), choose a stride $s\!\in\!\mathbb{N}$ and centers $\mathcal{C}=\{c_t\}_{t=1}^{T}$ with
\begin{equation}
  c_t \;=\; c_1 + s(t-1), \quad t=1,\dots,T.
  \label{eq:app-dinov3-center-prog}
\end{equation}
Define the tri-slice operator
\begin{equation}
  \mathrm{Tri}(X;c) \;=\; \operatorname{stack}_{k=-1}^{1}\, X_{\,c+k,\;:\,,\,:}
  \;\in\; \mathbb{R}^{3\times H\times W}.
  \label{eq:app-dinov3-tri-op}
\end{equation}
The per-study TRI sequence is
\begin{equation}
  \mathbf{X}^{\mathrm{TRI}}
  := \big(\mathrm{Tri}(X;c_t)\big)_{t=1}^{T}
  \in \mathbb{R}^{T\times 3\times H\times W}.
  \label{eq:app-dinov3-tri-seq}
\end{equation}

\paragraph{Preprocessing.}
Each TRI slice is resized to $S{\times}S$ and channelwise normalized (DINOv3/ImageNet). Let $\mathcal{P}_S$ denote this map. For a study and for a batch:
\begin{equation}
\label{eq:app-dinov3-prep}
\begin{aligned}
  \big(\mathcal{P}_S(\mathrm{Tri}(X;c_t))\big)_{t=1}^{T}
  &\in \mathbb{R}^{T\times 3\times S\times S},\\
  \mathbf{X}^{\mathrm{TRI}}_{S}
  &\in \mathbb{R}^{B\times T\times 3\times S\times S}.
\end{aligned}
\end{equation}

\paragraph{Patch tokenization and token lattice.}
Non-overlapping $P{\times}P$ patches on $S{\times}S$ yield a $g{\times}g$ grid with
\begin{equation}
  g \;=\; S/P, \qquad N \;=\; g^2 .
  \label{eq:app-dinov3-grid}
\end{equation}
We merge $(B,T)$ to run slices independently through the ViT, drop non-spatial tokens, and unmerge:
\begin{align}
  \widehat{\mathbf{X}}^{\mathrm{TRI}}_{S}
  &:= \operatorname{merge}_{(B,T)}\!\big(\mathbf{X}^{\mathrm{TRI}}_{S}\big)
  \in \mathbb{R}^{(BT)\times 3\times S\times S}, \label{eq:app-dinov3-merge}\\
  H &= \mathrm{ViT}\!\big(\widehat{\mathbf{X}}^{\mathrm{TRI}}_{S}\big)
  \in \mathbb{R}^{(BT)\times(1+N+R)\times d}, \label{eq:app-dinov3-vit}\\
  U &= H[:,\,1:1{+}N,\,:] \in \mathbb{R}^{(BT)\times N\times d}, \\
  \mathbf{U} &:= \operatorname{unmerge}_{(B,T)}(U)
  \in \mathbb{R}^{B\times T\times N\times d}. \label{eq:app-dinov3-unmerge}
\end{align}

Let the \emph{token lattice} be the index set
\[
\Omega \;=\; \{1,\dots,T\}\times\{1,\dots,N\},\qquad |\Omega|=TN.
\]
We flatten $(t,p)\!\mapsto\! i$ via a fixed bijection $\iota: \{1,\dots,|\Omega|\}\!\leftrightarrow\!\Omega$ and write
\[
u_i \;\equiv\; u_{t,p}\in\mathbb{R}^{d}\quad\text{for}\quad \iota(i)=(t,p).
\]
All aggregation heads operate \emph{only} on $\{u_i\}_{i\in\Omega}$, matching Sec.~\ref{sec:agg-overview}.

\subsection{Organ mask projection to tokens (flattened $i$)}
\label{app:dinov3:masks}

\paragraph{3D masks and per-slice extraction.}
Organ-group masks on the CT grid are
\begin{equation}
  M_o \in \{0,1\}^{D\times H\times W}, \qquad o\in\mathcal{O} .
  \label{eq:app-dinov3-mask-3d}
\end{equation}
For a TRI center $c_t$, extract the axial slice (after any mm-dilation/padding specified in App.~\Cref{tab:a1-ctrate-merges,tab:a2-merlin-merges}):
\begin{equation}
  \tilde{M}_{o,t} := M_o[c_t,\: :,\: :] \in \{0,1\}^{H\times W}.
  \label{eq:app-dinov3-mask-2d}
\end{equation}

\paragraph{Projection to the patch grid and flattening.}
Masks reuse the image geometry: NN resize to $S{\times}S$, then NN down to the $g{\times}g$ patch grid
\begin{equation}
  \Pi_{\mathrm{patch}}
  := \operatorname{NN}_{S\times S\!\to g\times g}
     \circ \operatorname{NN}_{H\times W\!\to S\times S},
  \label{eq:app-dinov3-Pi}
\end{equation}
yielding $\tilde{M}^{\mathrm{patch}}_{o,t}\!\in\!\{0,1\}^{g\times g}$. Flatten in the same raster order used by tokens to obtain binary indicators over the token lattice:
\begin{equation}
  m_{o,i}\in\{0,1\},\quad
  m_{o,i}=1 \;\Leftrightarrow\; \text{token } i\text{ lies in organ } o.
  \label{eq:app-dinov3-mask-flat}
\end{equation}
Let $\Omega_o=\{\,i\in\Omega:\, m_{o,i}=1\,\}$ denote the support at the token resolution.

\subsection{Aggregation modes with DINOv3 tokens (single index $i$)}
\label{app:dinov3:modes}

All heads here use only $\{u_i\}_{i\in\Omega}$. Per-location scorers are \emph{unary}
\begin{equation}
  s:\mathbb{R}^{d}\!\to\!\mathbb{R}\quad\text{(Linear $(d\!\to\!1)$ on tokens; no Q/K/V)}.
  \label{eq:app-dinov3-unary}
\end{equation}

\paragraph{(i) GAP baseline (uniform).}
\begin{align}
  h \;&=\; \tfrac{1}{|\Omega|}\sum_{i\in\Omega} u_i \in \mathbb{R}^{d}, \\
  z \;&=\; W^{\mathrm{global}} h + b^{\mathrm{global}} \in \mathbb{R}^{L}.
  \label{eq:app-dinov3-gap}
\end{align}

\paragraph{(ii) Global attention (mask-free).}
\begin{align}
  \alpha_i \;&=\; s_{\mathrm{glob}}(u_i), \\
  w_i \;&=\;
  \frac{\exp(\alpha_i/\tau)}{\sum_{j\in\Omega}\exp(\alpha_j/\tau)}, \quad \tau>0,\\
  h \;&=\; \sum_{i\in\Omega} w_i\,u_i, \qquad
  z \;=\; W^{\mathrm{global}} h + b^{\mathrm{global}} .
  \label{eq:app-dinov3-global}
\end{align}

\paragraph{(iii) Organ-masked attention.}
Each organ $o$ has a scorer $s_o$, temperature $\tau_o>0$, and optional priors $(\beta_o^{\mathrm{in}},\beta_o^{\mathrm{out}})$:
\begin{align}
  \tilde{\alpha}_{o,i}
  \;&=\; s_o(u_i)
  \;+\; \beta_o^{\mathrm{in}}\,m_{o,i}
  \;+\; \beta_o^{\mathrm{out}}(1-m_{o,i}), \label{eq:app-dinov3-masked-alpha}\\
  w_{o,i}
  \;&=\;
  \frac{\exp(\tilde{\alpha}_{o,i}/\tau_o)\, m_{o,i}}
       {\sum_{j\in\Omega_o}\exp(\tilde{\alpha}_{o,j}/\tau_o) + \varepsilon},
  \quad \varepsilon>0, \label{eq:app-dinov3-masked-softmax}\\
  h_o \;&=\; \sum_{i\in\Omega_o} w_{o,i}\,u_i, \qquad
  z_o \;=\; W^{(o)} h_o + b^{(o)} \in \mathbb{R}^{|\mathcal{L}_o|}.
  \label{eq:app-dinov3-masked-head}
\end{align}
If $\Omega_o=\varnothing$ at this resolution, we fall back to uniform pooling over $\Omega$ for that organ. Labels in $\mathcal{L}_{\texttt{other}}$ use the mask-free global head.

\paragraph{(iv) Organ-masked + scalar fusion (OSF).}
Compute per-organ scalars $u_o\in\mathbb{R}^{k}$ once from (dilated) masks (e.g., normalized volume, mean HU, border-contact flag). Let $b_o\!\in\!\{0,1\}$ be the border flag and $g_o=\sigma(\gamma_o)$ a learned gate:
\begin{align}
  \tilde{h}_o \;&=\; (1 - g_o\, b_o)\, h_o, \\
  \hat{h}_o \;&=\; [\,\tilde{h}_o;\,u_o\,] \in \mathbb{R}^{d+k}, \\
  z_o \;&=\; \mathrm{MLP}_o(\hat{h}_o) \in \mathbb{R}^{|\mathcal{L}_o|}.
  \label{eq:app-dinov3-osf}
\end{align}
Finally, assemble $z\in\mathbb{R}^{L}$ by stitching $\{z_o\}$ with the fixed label$\to$organ map.

\paragraph{Notes.}
(i) Temperatures $(\tau,\tau_o)$ and priors $(\beta_o^{\mathrm{in}},\beta_o^{\mathrm{out}})$ can be fixed or learned (we use learnable logits). (ii) \eqref{eq:app-dinov3-masked-softmax} reduces to global attention when $m_{o,i}\!\equiv\!1$. (iii) The same equations apply to voxel backbones by reinterpreting $i$ as a flattened voxel index; only mask--to--lattice alignment changes.